\documentclass{article} 
\usepackage{iclr2027_conference,times}

\usepackage{amsmath,amsfonts,bm}

\def\eqref#1{equation~\ref{#1}}

\def\1{\bm{1}}

\DeclareMathAlphabet{\mathsfit}{\encodingdefault}{\sfdefault}{m}{sl}
\SetMathAlphabet{\mathsfit}{bold}{\encodingdefault}{\sfdefault}{bx}{n}

\definecolor{linkpurple}{RGB}{255,0,0}
\definecolor{SkyBlue}{RGB}{135,206,235}
\definecolor{urlorange}{RGB}{190,90,20}
\usepackage[colorlinks=true, linkcolor=linkpurple, citecolor=SkyBlue, urlcolor=urlorange]{hyperref}
\usepackage{url}

\usepackage{graphicx}
\usepackage{placeins}
\newcommand{\NA}{\textcolor{black!35}{--}}
\newcommand{\toprule}{\noalign{\hrule height .8pt\vskip 2pt}}
\newcommand{\midrule}{\noalign{\vskip 1pt\hrule height .4pt\vskip 1pt}}
\newcommand{\bottomrule}{\noalign{\vskip 1pt\hrule height .8pt}}

\title{MemoWM: How World Models Change What Agents Need to Remember}

\author{%
\begin{tabular}{c}
Bingfan Zeng$^{1,*}$ \quad
Zhisheng Chen$^{2,*}$ \quad
Chenbo Sang$^{3,*}$ \quad
Zhengwei Xie$^{4}$ \quad
Jinpeng Wang$^{5}$ \\
Xiangchen Guan$^{6}$ \quad
Rui Qian$^{7}$ \quad
Zheng Lu$^{6,\dagger}$ \quad
Jingwei Song$^{8,\dagger}$ \\[6pt]
\normalfont\small $^{1}$South China University of Technology \quad
\normalfont\small $^{2}$University of the Chinese Academy of Sciences \\
\normalfont\small $^{3}$Beijing University of Posts and Telecommunications \quad
\normalfont\small $^{4}$University of Science and Technology of China \\
\normalfont\small $^{5}$Tsinghua University \quad
\normalfont\small $^{6}$Peking University \quad
\normalfont\small $^{7}$Fudan University \quad
\normalfont\small $^{8}$Shanghai Jiao Tong University \\[5pt]
\normalfont\small $^{*}$Equal contribution \qquad
$^{\dagger}$Corresponding authors \\[3pt]
\normalfont\small\texttt{202430551902@mail.scut.edu.cn,}
\texttt{zhisheng.researcher@gmail.com}
\end{tabular}%
}

\iclrfinalcopy 
\begin{document}

\maketitle
\vspace{-0.22in}

\begin{abstract}
Long-term agents face growing storage demands as they accumulate experience. World models capture reusable regularities that can reduce the information stored for each experience. We formulate the problem of memory allocation conditioned on a world model and introduce MemoWM, a framework that uses shared predictions to compress retained information and reconstruct omitted content. Its task-aware allocation rule balances the expected impact of reconstruction errors against storage cost, retaining information with downstream value beyond the predictive prior. Across five long-term agent-memory benchmarks, MemoWM achieves 42.42\% average answer accuracy, exceeding the strongest baseline by 2.62 percentage points, while reducing average experience-specific storage by 53.9\% relative to MIRIX, the most storage-efficient baseline. Further analysis shows that stronger world models reduce per-experience storage at comparable task quality. Accounting for model parameters reveals a trade-off between shared model capacity and recurring storage costs, with the capacity that minimizes total storage increasing as more interactions are retained. Our code is available at
\url{https://github.com/Feld-maxiu/MemoWM}.
\end{abstract}

\vspace{-1.2em}
\begin{figure}[h]
\centering
\includegraphics[width=0.9\linewidth,trim=20 10 15 22,clip]{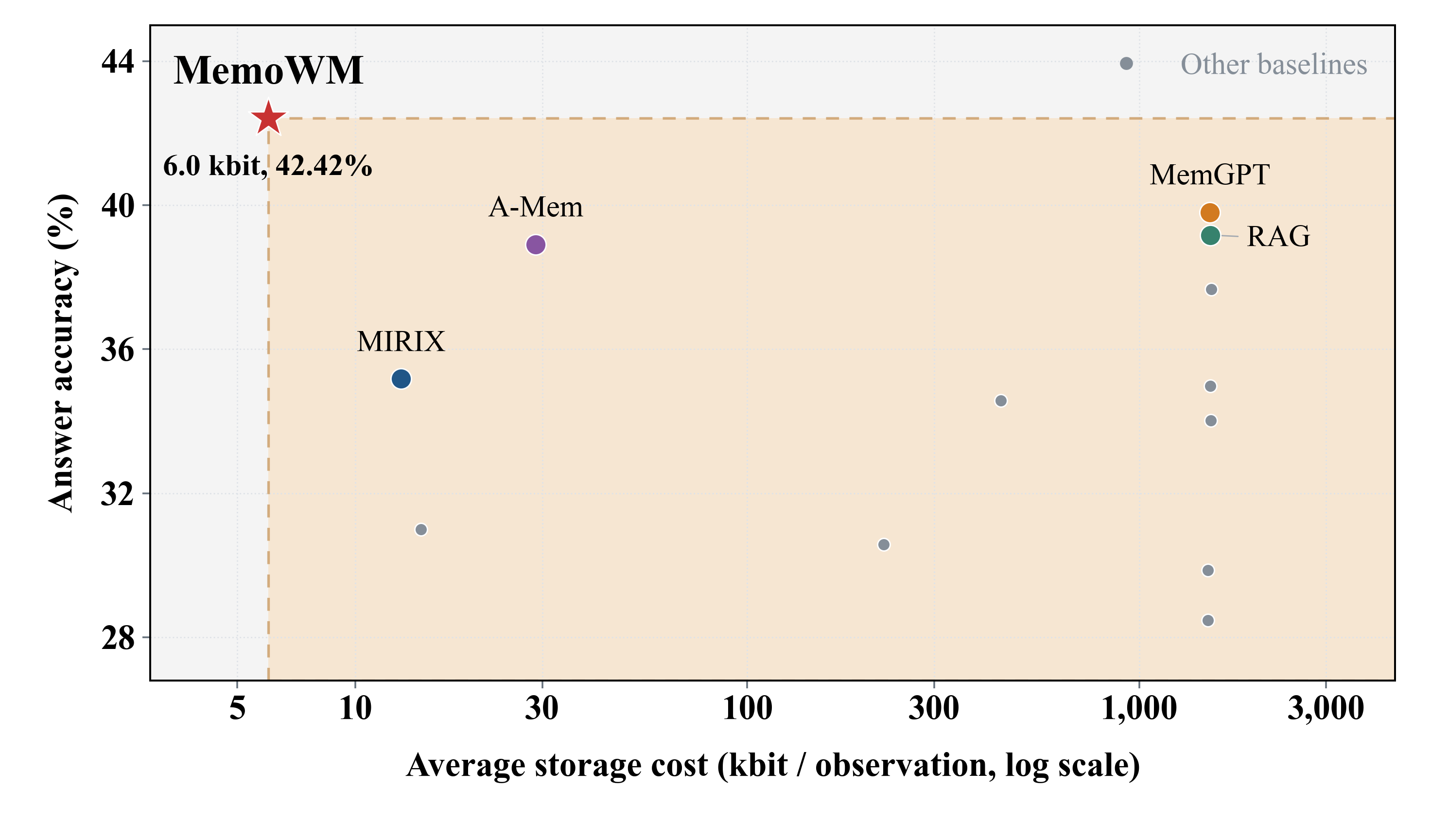}
\vspace{-0.2em}
\caption{Average accuracy versus experience-specific storage.}
\label{fig:abstract-tradeoff}
\end{figure}
\vspace{-1em}

\section{Introduction}\label{sec:introduction}

Agent research is moving from answering isolated questions to acting
continuously over time. An agent that can genuinely carry out a task must
repeatedly observe its environment, take actions, revise its plan, and bring
what happened before into its next decision---over hours, days, or longer
\citep{yao2022react,deng2023mind2web,zhou2024webarena}. GUI and web
agents make this demand especially concrete: they execute long sequences of
actions through websites and computer interfaces whose state persists across
steps \citep{he2024webvoyager,xie2024osworld}. For such
agents, memory is not merely extra context. It is what gives behavior
continuity and allows experience to accumulate.

Giving an agent a past, however, does not require preserving that past
verbatim. Long interactions repeatedly expose the same page structure,
stable control layouts, persistent environmental regularities, and future
content that can be inferred from earlier states and actions \citep{wu2024longmemeval,rasmussen2025zep}. What matters
for a later decision may be only whether a click succeeded, an input was
accepted, or a small part of the state changed. Writing every observation in
full makes persistent storage grow continually with the agent's lifetime.
Storing only what has the largest prediction error is not enough either: a
nearly unchanged fact can still be decisive for the task, while an
unpredictable interface change may be irrelevant noise. This tension is
largely left unresolved by memory systems that gate or consolidate experience
using novelty, surprise, prediction error, or downstream utility
\citep{wang2023voyager,tan2025prospect,xiong2026memory}.

This leads to a more basic question for long-term memory: \textbf{once an
agent knows what usually happens in its environment, how much of a particular
experience must still be stored explicitly?} Answering it requires separating
two forms of persistent information. A world model internalizes regularities
that can be reused across experiences, whereas episodic memory pays a new
storage cost for each event that actually occurred \citep{bruce2024genie,zhou2024dino,assran2025v}. When recall has access to
the same world model, content recoverable from shared knowledge need not be
stored repeatedly in every episode. Episodic memory should instead carry what
cannot be recovered from the existing prediction and still matters for future
reasoning or action. Recoverability alone, however, does not determine what
deserves storage: the decision must jointly account for whether information
can be predicted and whether an error would affect the task. The former is a
property of the world model; the latter is a property of the information's
task value \citep{bar2025navigation}.

We therefore adopt a direct principle: \textbf{leave what the world model can
explain to the model, and store in episodic memory only what cannot be safely
recovered and matters to future tasks}. Under this view, memory is no longer
just a container for observations, but a division of labor between shared
knowledge and individual experience. A world model shared by the writer and
decoder becomes a predictive prior: it supplies both a coding distribution for
information that must be preserved exactly and a reconstruction for
information that may be omitted. Prediction is not only a signal for
deciding what to remember; it becomes part of the memory representation
itself.

We instantiate this principle in \textbf{MemoWM}, an episodic-memory
framework that uses a world model as a shared predictive prior. Each
observation is first encoded into a compact, query-independent state so that
the same memory can serve future questions unknown at write time. The state
is discretized, and an action-conditioned world model predicts a conditional
distribution for every coordinate from reconstructed history. Offline
single-coordinate counterfactual readouts estimate how much an error at each
coordinate affects task answers. At write time, MemoWM combines this
task-impact prior with the current prediction risk and expected coding cost to
decide which true values must be stored and which can be reconstructed by the
world model. Because the decision depends only on the shared predictive
distribution, frozen task statistics, and a fixed threshold, it requires
neither future queries nor a transmitted state-specific mask. Retained true
codes and model-filled coordinates form the reconstructed state, which is fed
back as history for the next prediction, yielding a synchronized closed-loop
memory.

Across five long-term agent-memory benchmarks, MemoWM achieves the highest
average answer accuracy ($42.42\%$, Table~\ref{tab:main-results}) while using
the lowest experience-specific storage on every benchmark: it exceeds the
strongest baseline by $2.62$ points and reduces average storage by $53.9\%$
relative to MIRIX. Further
analysis shows that stronger world models internalize more reusable structure
and thereby reduce the storage required for each experience. Once the model's
own parameters are also counted, however, the storage-optimal capacity depends
on how many interactions must be remembered and shifts toward larger models as
the agent's lifetime grows.

Our contributions are threefold.

\textbf{We recast long-term agent memory around the question of how
much information an experience must retain beyond existing world knowledge.}
It distinguishes selecting which experiences matter from
determining how many bits a selected experience still needs, and unifies world
models and episodic memory as persistent information with different
amortization costs.

\textbf{We introduce \textbf{MemoWM}, which turns world-model
predictions into a task-usable prior for episodic memory.} It combines a
query-independent state interface, a shared predictive prior, and offline task
impact estimates to retain only important information that cannot be recovered, while the allocation rule and closed-loop
reconstruction keep writing and decoding synchronized.

\textbf{We characterize the trade-off among world-model capacity,
episodic storage, and task quality.} Stronger predictors lower the recurring
cost of storing each experience, but the model itself incurs a
parametric cost, and the storage-optimal world-model capacity grows
with the memory horizon.

\section{Related Work}
\label{sec:related-work}

\paragraph{World models and memory.}
World models learn environment dynamics for future-state prediction and
planning. Existing connections with memory largely focus on improving
these capabilities. R2I incorporates long-range state-space memory into
world models for reinforcement learning. R-WoM grounds simulations in
retrieved external knowledge, while MemWM and WorldEvolver use transition
rules, past states, and prediction failures to improve simulation and
planning \citep{mei2026r,wang2026memwm,zhang2026self}. MemoWM studies the complementary direction: how predictions
from an available world model change the information needed to retain an
experience.

\paragraph{Selective and compressed agent memory.}
Agent-memory systems use importance and temporal decay to prioritize
records \citep{omri2026agent,zhong2024memorybank}, structured extraction and
consolidation to compact their contents \citep{chhikara2025mem0,liu2026simplemem}, and evolving
links or typed stores to organize knowledge \citep{NEURIPS2025_19909c36,wang2025mirix}. Closely related
work also uses prediction to guide memory construction: NEMORI distills
information that existing knowledge fails to anticipate, while D-MEM
uses reward prediction error to route inputs between inexpensive handling
and memory evolution \citep{nan2025nemori,song2026d}. In these approaches, prediction informs distillation
or update decisions. MemoWM additionally makes the predictive distribution
available during decoding, so it determines both conditional coding costs
and how omitted content is reconstructed. The resulting allocation is
coordinate-level and task-aware: prediction error probability and the
task impact of an error are separately, then balanced against
expected coding cost. 

\paragraph{Predictive and conditional compression.}
Predictive coding represents information unexplained by predictions \citep{singh2026nous}.
Scale-hyperprior image compression and deep contextual video compression
further exploit statistical or temporal context for conditional coding.
Rate--distortion theory formalizes coding with side information, including
the Wyner-Ziv \citep{aaron2002wyner} setting where it is available only at the decoder; MemoWM
uses a predictive prior shared by writer and decoder. In persistent agent
memory, the distortion of interest is the effect of an omission on future
answers. MemoWM couples this task criterion with a retention rule that
both sides reproduce from shared reconstructed history, allowing selective
omission without a per-state mask. This connects conditional compression
to task-aware memory allocation while preserving the synchronized
reconstruction needed to decode successive states.

\section{MemoWM}\label{memowm}

\begin{figure*}[t]
\centering
\begin{minipage}[t]{\textwidth}
\centering
\includegraphics[width=\linewidth]{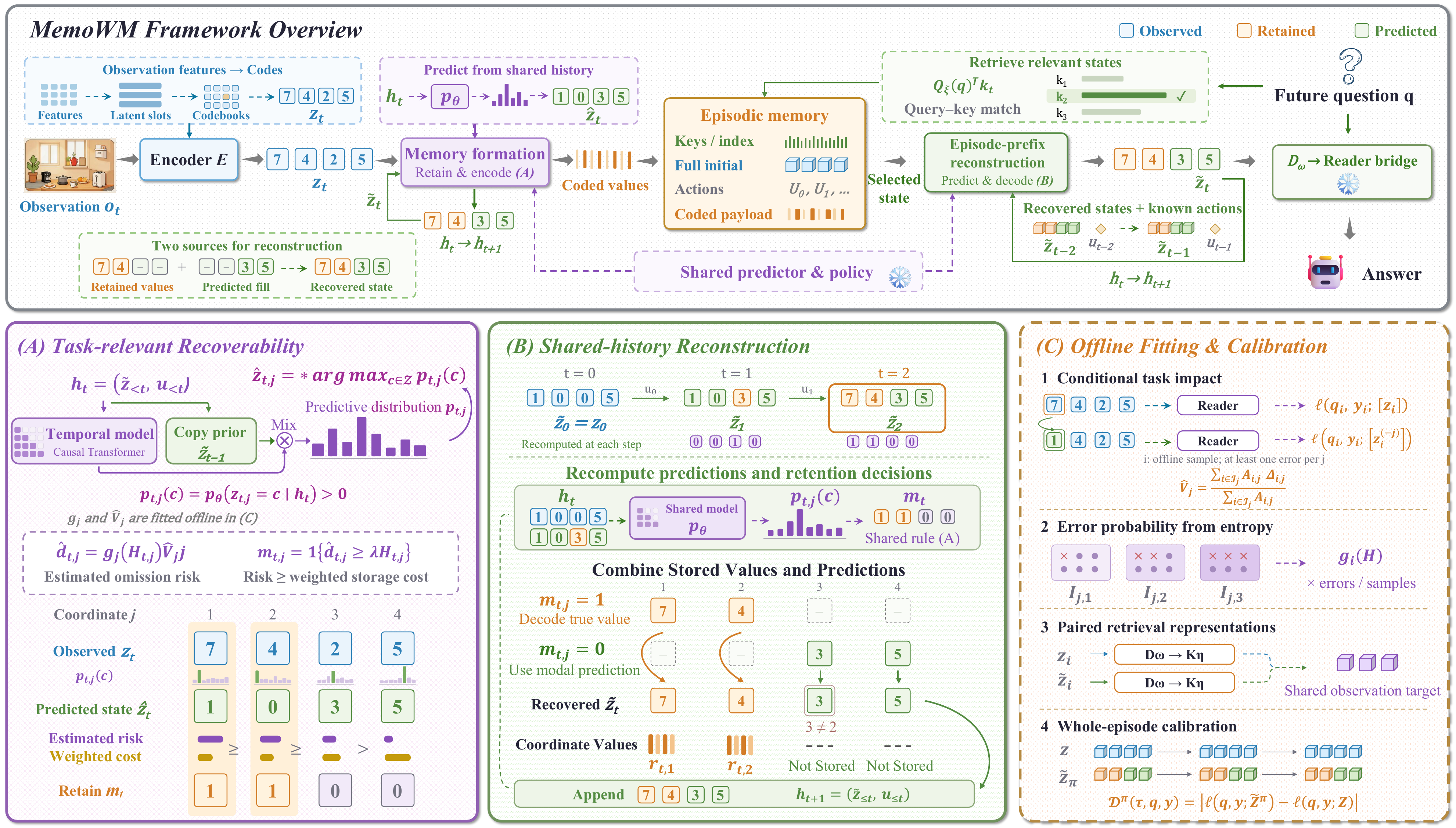}
\caption{\textbf{Overview of the MemoWM framework.} Coordinates are retained when their estimated task impact exceeds the weighted storage cost; others are filled by a causal world model. (A) Task-relevant retention, (B) shared-history reconstruction, and (C) offline fitting and calibration. Reconstructed states are retrieved through the Reader Bridge to answer future queries.}
\label{fig:framework}
\end{minipage}
\end{figure*}

As shown in Figure \ref{fig:framework}, MemoWM models long-term memory as joint coding by a shared world model
and an episodic store. Each observation is mapped to a query-independent
discrete state. An action-conditioned predictor estimates each
coordinate from the reconstructed history. A coordinate is retained when
the expected task impact of omitting it exceeds its predicted storage
cost; otherwise, the model prediction fills it. Because the writer and
decoder use the same reconstructed history, no retention mask is
transmitted for each state. Task-impact statistics, entropy calibration,
retrieval representations, and the retention threshold are fitted
offline and fixed before deployment.

\subsection{Problem Definition}\label{problem-definition}

Consider a trajectory \(\tau=(o_0,u_0,\ldots,u_{T-1},o_T)\), where
\(u_t\) follows observation \(o_t\). Each observation is encoded as
\(z_t=E(o_t)\in\mathcal Z^J\), where \(\mathcal Z=\{0,\ldots,C-1\}\) and
\(C\ge2\). Memory formation does not require future questions, so one
memory can support different tasks. Let \(\Phi\) denote the shared world
model and other shared parameters. Under retention policy \(\pi\), the
full state sequence \(Z=(z_0,\ldots,z_T)\) is reconstructed as
\(\widetilde Z^\pi=(\tilde z_0,\ldots,\tilde z_T)\).

For a future question \(q\) and reference answer \(y\), let
\(P_R(y\mid q,Z)\) be the probability assigned by the fixed retrieval
and reading procedure. Define the answer loss
\(\ell(q,y;Z)=-\log_2P_R(y\mid q,Z)\) and the distortion caused by
omission as

\[
\begin{aligned}
\mathcal D^\pi(\tau,q,y)
&=\left|\ell(q,y;\widetilde Z^\pi)-\ell(q,y;Z)\right|,\\
\min_{\pi\in\Pi_{\mathrm{shared}}}\quad
&\mathbb E_\tau[R_{\mathrm{epi}}^\pi(\tau)]
\quad\mathrm{s.t.}\quad
\mathbb E_{\tau,q,y}[\mathcal D^\pi(\tau,q,y)]\le\epsilon .
\end{aligned}
\tag{1}
\]

Here \(\epsilon\ge0\) is the allowed average change in answer loss.
Policies in \(\Pi_{\mathrm{shared}}\) use only information available to
both the writer and decoder. \(R_{\mathrm{epi}}^\pi\) is the additional
storage for one episode, while shared parameters incur a separate cost
\(B_{\mathrm{shared}}(\Phi)\) that can be amortized across episodes. We
fix \(\Phi\) below and describe retention, reconstruction, and reading.

\subsection{Task-Relevant Recoverability}\label{task-relevant-recoverability}

For each coordinate, the retention decision combines the probability of
a prediction error, its effect on the answer, and its storage cost. For
\(t\ge1\), the world model conditions on reconstructed states and known
actions, \(h_t=(\tilde z_{<t},u_{<t})\), and produces

\[
p_{t,j}(c)=p_\theta(z_{t,j}=c\mid h_t)>0.
\]

This distribution both codes retained values and fills omitted coordinates with the modal prediction \(\hat z_{t,j}=\arg\max_{c\in\mathcal Z}p_{t,j}(c)\). This prediction is shared by writer and decoder. The ideal code length of the true value and its expected length under the model distribution are

\[
r_{t,j}=-\log_2p_{t,j}(z_{t,j}),\qquad
H_{t,j}=-\sum_{c\in\mathcal Z}p_{t,j}(c)\log_2p_{t,j}(c).
\tag{2}
\]

To measure the effect of an error in coordinate \(j\), replace
\(z_{t,j}\) by \(\hat z_{t,j}\) and denote the resulting state by
\(z_t^{(-j)}\). For the same question and answer, define

\[
\Delta_{t,j}
=\left|\ell(q,y;[z_t^{(-j)}])-\ell(q,y;[z_t])\right|,
\qquad
A_{t,j}=\mathbf 1\{\hat z_{t,j}\ne z_{t,j}\}.
\tag{3}
\]

The brackets denote reading from the state alone. Since
\(\Delta_{t,j}=0\) when the prediction is correct,

\[
\underbrace{\mathbb E_{\mathrm{obs}}[\Delta_{t,j}\mid h_t]}_{\text{local omission risk}}
=
\underbrace{\Pr_{\mathrm{obs}}(A_{t,j}=1\mid h_t)}_{\text{error probability}}
\underbrace{\mathbb E_{\mathrm{obs}}[\Delta_{t,j}\mid A_{t,j}=1,h_t]}_{\text{task impact given an error}}.
\tag{4}
\]

Offline data provide a coordinate-level error impact \(\widehat V_j\)
and an entropy-to-error mapping \(g_j(H)\). Online memory formation uses
\(\hat d_{t,j}=g_j(H_{t,j})\widehat V_j\) as the estimated omission
risk. These quantities are shared and fixed, so the decision requires
neither future questions nor a reader call.

Let \(m_{t,j}=1\) indicate retention and let \(\lambda\ge0\) control the
storage--distortion trade-off. The coordinate-wise proxy objective is

\[
\begin{aligned}
\widehat{\mathcal J}_t(m)
&=\sum_{j=1}^{J}\left[(1-m_j)\hat d_{t,j}
+\lambda m_jH_{t,j}\right],\qquad m\in\{0,1\}^J,\\
m_{t,j}
&=\mathbf 1\!\left\{\hat d_{t,j}\ge\lambda H_{t,j}\right\}.
\end{aligned}
\tag{5}
\]

Ties retain the true value. \(H_{t,j}\) is used only to estimate storage
cost; the actual coded rate uses \(r_{t,j}\). Equation (5) optimizes a
coordinate-wise proxy, while joint omissions can interact through
retrieval and future predictions. The threshold \(\lambda\) is 
selected by full-episode calibration in Section \ref{offline-fitting-and-calibration}.

\subsection{Shared-History Reconstruction}\label{shared-history-reconstruction}

Omitted coordinates are filled with predictions and fed into
predictions. The writer and decoder use the same history. The
initial state is stored in full, \(\tilde z_0=z_0\), and later states
are reconstructed as

\[
\tilde z_{t,j}=
\begin{cases}
z_{t,j},&m_{t,j}=1,\\
\hat z_{t,j},&m_{t,j}=0.
\end{cases}
\tag{6}
\]

With identical initial states, actions, shared parameters, and
computation rules, both sides obtain identical predictions and retention
decisions. The decoder can therefore determine which values to read
without receiving a per-state mask. The action \(u_{t-1}\) must be
available before reconstructing \(z_t\). If each coordinate of the
initial state uses \(\lceil\log_2C\rceil\) bits, episodic storage is

\[
R_{\mathrm{epi}}^\pi(\tau)
=J\lceil\log_2C\rceil
+\sum_{t=1}^{T}\sum_{j=1}^{J}m_{t,j}r_{t,j}
+B_{\mathrm{aux}}^\pi(\tau).
\tag{7}
\label{eq:experience-storage}
\]

Here \(B_{\mathrm{aux}}^\pi\) covers actions and other per-episode
auxiliary content that cannot be recovered losslessly from counted
content; enabled symbol anchors are included. Regenerable retrieval
vectors and indices are excluded from the main table's logical memory
load; their implementation overhead may be reported separately if they
are stored as independent persistent state. Equation (7) uses ideal code
lengths and excludes file overhead. For a future question, retrieval
operates on reconstructed states. State training uses paired full and
reconstructed views from the same episode, while questions provide
supervision rather than inputs to the observation encoder.

\subsection{Offline Fitting and Calibration}\label{offline-fitting-and-calibration}

Offline fitting produces the shared configuration used by the writer and
decoder. It estimates task impact and prediction reliability, trains a
retrieval head that is robust to reconstruction, and selects the
retention threshold.

\begin{table*}[t]
\caption{Main results across five long-horizon memory benchmarks. (a) The
upper table reports answer accuracy (\%); (b) the middle table reports the
experience-specific storage rate (kbit/observation); (c) the break-even point
is the number of retained observations at which MemoWM's total storage
is no greater than the listed comparator's.
LME-V2 denotes LongMemEval-V2, Omni-SM denotes Omni-SimpleMem, and Avg. is the
unweighted mean across benchmarks.}
\label{tab:main-results}
\centering
\small
\setlength{\tabcolsep}{4pt}
\renewcommand{\arraystretch}{0.95}
\begin{tabular*}{\textwidth}{@{\extracolsep{\fill}}l*{6}{r}@{}}
\toprule
\textbf{Method} & \textbf{WorldMemArena} & \textbf{AMA-Bench}
& \textbf{LME-V2} & \textbf{WebQuest} & \textbf{MemEye} & \textbf{Avg.} \\
\midrule
RAG & 52.97{\scriptsize$\pm$0.12} & \textbf{44.87}{\scriptsize$\pm$0.70} & 40.25{\scriptsize$\pm$0.69} & 22.35{\scriptsize$\pm$0.50} & 35.42{\scriptsize$\pm$0.93} & 39.17 \\
UniversalRAG & 40.38{\scriptsize$\pm$0.19} & 40.35{\scriptsize$\pm$0.88} & 34.56{\scriptsize$\pm$1.18} & 21.68{\scriptsize$\pm$0.75} & \underline{37.95}{\scriptsize$\pm$1.24} & 34.98 \\
MEM1 & 45.66{\scriptsize$\pm$0.31} & 14.96{\scriptsize$\pm$1.64} & 29.41{\scriptsize$\pm$1.83} & 20.25{\scriptsize$\pm$1.01} & 32.05{\scriptsize$\pm$1.93} & 28.47 \\
MemAgent & 43.03{\scriptsize$\pm$0.37} & 29.18{\scriptsize$\pm$1.60} & 29.33{\scriptsize$\pm$1.91} & 16.90{\scriptsize$\pm$0.78} & 30.88{\scriptsize$\pm$2.34} & 29.86 \\
A-Mem & 55.37{\scriptsize$\pm$0.28} & 34.85{\scriptsize$\pm$1.31} & 41.28{\scriptsize$\pm$1.48} & \underline{26.70}{\scriptsize$\pm$0.85} & 36.33{\scriptsize$\pm$1.28} & 38.91 \\
MemGPT & \underline{58.43}{\scriptsize$\pm$0.20} & 35.44{\scriptsize$\pm$0.81} & \textbf{44.97}{\scriptsize$\pm$1.29} & 24.55{\scriptsize$\pm$0.87} & 35.61{\scriptsize$\pm$1.16} & \underline{39.80} \\
SimpleMem & 41.32{\scriptsize$\pm$0.22} & 19.26{\scriptsize$\pm$1.40} & 31.56{\scriptsize$\pm$1.32} & 26.00{\scriptsize$\pm$0.65} & 36.88{\scriptsize$\pm$1.20} & 31.00 \\
Omni-SM & 42.54{\scriptsize$\pm$0.32} & 27.56{\scriptsize$\pm$1.50} & 38.12{\scriptsize$\pm$1.94} & 25.55{\scriptsize$\pm$0.72} & \textbf{39.15}{\scriptsize$\pm$1.51} & 34.58 \\
M2A & 50.35{\scriptsize$\pm$0.18} & 42.61{\scriptsize$\pm$1.02} & 40.18{\scriptsize$\pm$1.05} & 25.45{\scriptsize$\pm$0.99} & 29.72{\scriptsize$\pm$2.20} & 37.66 \\
ViLoMem & 49.59{\scriptsize$\pm$0.39} & 24.19{\scriptsize$\pm$2.00} & 34.79{\scriptsize$\pm$2.27} & 15.65{\scriptsize$\pm$0.91} & 28.65{\scriptsize$\pm$2.28} & 30.57 \\
MIRIX & 45.63{\scriptsize$\pm$0.43} & 37.24{\scriptsize$\pm$1.88} & 39.62{\scriptsize$\pm$2.01} & 21.30{\scriptsize$\pm$0.98} & 32.10{\scriptsize$\pm$2.05} & 35.18 \\
AUGUSTUS & 39.54{\scriptsize$\pm$0.34} & 38.45{\scriptsize$\pm$1.35} & 32.15{\scriptsize$\pm$1.88} & 22.73{\scriptsize$\pm$0.65} & 37.21{\scriptsize$\pm$1.71} & 34.02 \\
\midrule
\textbf{MemoWM} & \textbf{60.18}{\scriptsize$\pm$0.23} & \underline{43.58}{\scriptsize$\pm$0.85} & \underline{42.91}{\scriptsize$\pm$1.55} & \textbf{27.85}{\scriptsize$\pm$0.57} & 37.58{\scriptsize$\pm$0.99} & \textbf{42.42} \\
\bottomrule
\end{tabular*}
\vspace{1pt}

\begin{tabular*}{\textwidth}{@{\extracolsep{\fill}}l*{6}{r}@{}}
\toprule
\textbf{Method} & \textbf{WorldMemArena} & \textbf{AMA-Bench}
& \textbf{LME-V2} & \textbf{WebQuest} & \textbf{MemEye} & \textbf{Avg.} \\
\midrule
RAG & 1809.3 & 5.6 & 71.9 & 4141.2 & 1560.5 & 1517.7 \\
UniversalRAG & 1805.4 & 5.7 & 71.3 & 4142.1 & 1573.7 & 1519.6 \\
MEM1 & 1804.3 & 5.6 & 18.1 & 4140.1 & 1523.1 & 1498.2 \\
MemAgent & 1803.5 & 5.7 & 25.8 & 4140.2 & 1519.2 & 1498.9 \\
A-Mem & 12.1 & 5.9 & 78.3 & 24.8 & 23.4 & 28.9 \\
MemGPT & 1809.0 & 5.1 & 74.9 & 4153.6 & 1536.5 & 1515.8 \\
SimpleMem & 9.8 & 3.5 & 12.0 & 32.5 & \underline{15.8} & 14.7 \\
Omni-SM & 331.3 & 3.6 & 55.2 & 648.0 & 1182.7 & 444.2 \\
M2A & 1814.8 & 6.1 & 80.6 & 4167.0 & 1559.1 & 1525.5 \\
ViLoMem & 15.0 & 4.3 & 11.1 & 38.5 & 1047.3 & 223.2 \\
MIRIX & \underline{8.2} & \underline{3.2} & \underline{10.8} & \underline{21.0} & 22.1 & \underline{13.1} \\
AUGUSTUS & 1813.7 & 5.7 & 79.4 & 4185.0 & 1540.2 & 1524.8 \\
\midrule
\textbf{MemoWM} & \textbf{4.3} & \textbf{2.3} & \textbf{9.4} & \textbf{10.3} & \textbf{3.8} & \textbf{6.0} \\
\bottomrule
\end{tabular*}
\par\vspace{1pt}
\begin{tabular*}{\textwidth}{@{\extracolsep{\fill}}l*{5}{r}@{}}
\toprule
\textbf{Metric} & \textbf{WorldMemArena} & \textbf{AMA-Bench}
& \textbf{LME-V2} & \textbf{WebQuest} & \textbf{MemEye} \\
\midrule
Comparator & MIRIX & MIRIX & MIRIX & MIRIX & SimpleMem \\
Break-even point (k obs.) & 119.4 & 517.2 & 332.5 & 43.5 & 38.8 \\
\bottomrule
\end{tabular*}
\end{table*}

For each coordinate, offline state-question-answer samples are used for
a single-coordinate counterfactual measurement, yielding
\(\widehat V_j\) from examples with prediction errors. Under the
full-history reference policy, predictive entropy and error indicators
are collected in entropy bins to form \(g_j(H)\in[0,1]\): populated bins
use a pooled error frequency, while empty bins use the coordinate
marginal rate. For retrieval training, the full and recursively
reconstructed states of the same observation are mapped through
\(D_\omega\) and the retrieval head \(K_\eta\). Contrastive and
consistency objectives align both views with a teacher representation;
\(K_\eta\) is then frozen.

Finally, a finite candidate set \(\Lambda\) is evaluated on independent
calibration episodes. For each \(\lambda\in\Lambda\), the system
reconstructs complete episodes from their full initial states using the
retention rule and shared-history recursion, then evaluates the fixed
retrieval and reading procedure. The candidate with minimum episodic
storage subject to the constraint in Eq. (1) is selected and fixed for
evaluation. This full-episode calibration accounts for interactions
among omissions, retrieval changes, and later prediction histories. The
training order is observation representation and reader interface,
quantizer and world model, offline risk calibration, retrieval-head
training, and threshold selection.

\providecommand{\NA}{\textcolor{black!35}{--}}
\providecommand{\toprule}{\noalign{\hrule height .8pt\vskip 2pt}}
\providecommand{\midrule}{\noalign{\vskip 1pt\hrule height .4pt\vskip 1pt}}
\providecommand{\bottomrule}{\noalign{\vskip 1pt\hrule height .8pt}}

\section{Experiments}\label{sec:experiments}
\FloatBarrier

\subsection{Experimental Setup}
\label{sec:experimental-setup}

\paragraph{Benchmarks.}
We use five long-term agent-memory benchmarks in web- or GUI-based settings:
\textbf{AMA-Bench}\citep{zhao2026ama} (recall, causal inference, state updating,
and abstraction; \emph{Web Task Execution} subset), \textbf{LongMemEval-V2
(LME-V2)}\citep{wu2026longmemeval} (environment knowledge accumulated across
trajectories; \emph{Small} setting), \textbf{WorldMemArena}
\citep{liu2026worldmemarena} (multimodal memory over interaction histories;
\emph{Web} subset), \textbf{WebQuest}\citep{wang2024webquest} (multimodal
reasoning over webpage sequences; \emph{Trace QA}), and
\textbf{MemEye}\citep{guo2026memeye} (visual-centric multimodal memory).
Data construction and overlap checks are in
\hyperref[app:webarena-data]{Appendix~\ref*{app:webarena-data}}.

\paragraph{Baselines.}
We compare against twelve baselines spanning retrieval (RAG
\citep{lewis2020retrieval}, UniversalRAG \citep{yeo2026universalrag}),
context-management and recurrent memory (MemGPT \citep{packer2023memgpt}, MEM1
\citep{zhou2026mem1}, MemAgent \citep{yu2026memagent}), structured extraction
(A-Mem \citep{NEURIPS2025_19909c36}, SimpleMem \citep{liu2026simplemem}, MIRIX
\citep{wang2025mirix}), and multimodal memory (Omni-SimpleMem
\citep{liu2026omni}, M2A \citep{feng2026m2a}, ViLoMem \citep{bo2025agentic},
AUGUSTUS \citep{jain2025augustus}). All methods share the same backbone,
reader, and LLM judge per benchmark protocol.

\paragraph{Evaluation protocol.}
WorldMemArena and LME-V2 use Qwen3.5-9B as reader and judge, and AMA-Bench
uses Qwen3-32B under its open-ended QA protocol; WebQuest and MemEye follow
their official protocols. We report each benchmark's native answer accuracy,
averaged over four random seeds with standard deviation. Details are given in
\hyperref[app:evaluation-settings]{Appendix~\ref*{app:evaluation-settings}}.

\paragraph{Storage cost.}
We report experience-specific persistent storage in kbit/observation, counting
retained content and observation-specific metadata while excluding shared model
weights, which are amortized separately over retained experience. Unified
accounting rules are given in
\hyperref[app:storage-accounting]{Appendix~\ref*{app:storage-accounting}}.

\subsection{Main Results}
\label{sec:main-results}

\noindent\textbf{MemoWM achieves the strongest overall accuracy--storage
trade-off across the five benchmarks.}
It obtains the highest mean answer accuracy ($42.42\%$,
Table~\ref{tab:main-results}) while using the lowest experience-specific
storage on every benchmark. Relative to MIRIX, the most storage-efficient
baseline on average, MemoWM reduces mean storage by $53.9\%$ while improving
accuracy by $7.24$ points. Even where higher-accuracy alternatives exist, on
AMA-Bench, LME-V2, and MemEye, MemoWM uses $58.9$--$99.7\%$ less storage for
only $1.29$--$2.06$ points lower accuracy. These results motivate our next
question: where do these storage savings come from?

\FloatBarrier

\subsection{A Composable Principle for Memory Systems}
\label{sec:component-ablation}

\begin{table}[!ht]
\noindent
\begin{minipage}[t]{0.36\textwidth}
\vspace{0pt}
\raggedright
\textbf{Beyond quantization.} Quantization provides a compact state
representation, but does not determine how much of that state must be stored
explicitly. Table~\ref{tab:component-ablation} separates the gains from
representation compression, predictive coding, and task-aware omission.
With the quantizer fixed, world-model coding saves bits on predictable
coordinates.
\end{minipage}\hfill
\begin{minipage}[t]{0.62\textwidth}
\vspace{-\abovecaptionskip}
\caption{Ablation and M2A integration. AC: accuracy (\%);
Rate: experience-specific storage (kbit/observation).}
\label{tab:component-ablation}
\centering
\small
\setlength{\tabcolsep}{4pt}
\renewcommand{\arraystretch}{1.08}
\begin{tabular*}{\linewidth}{@{\extracolsep{\fill}}lrrrrrr@{}}
\toprule
& \multicolumn{2}{c}{WMA} & \multicolumn{2}{c}{AMA-Bench}
& \multicolumn{2}{c}{LME-V2} \\
\cline{2-3}\cline{4-5}\cline{6-7}
Variant & AC & Rate & AC & Rate & AC & Rate \\
\midrule
No quantizer & 60.33 & 262.5 & 43.75 & 263.0 & 43.04 & 272.2 \\
Quantizer only & 60.27 & 6.2 & 43.27 & 7.1 & 42.98 & 14.3 \\
WM, all-send & 60.27 & 5.6 & 43.27 & 3.5 & 42.98 & 11.2 \\
\textbf{MemoWM} & 60.18 & \textbf{4.3} & 43.58 & \textbf{2.3} & 42.91 & \textbf{9.4} \\
\midrule
M2A & 50.35 & 1814.8 & 42.61 & 6.1 & 40.18 & 80.6 \\
M2A + MemoWM & 50.49 & 19.2 & 42.53 & 3.1 & 40.53 & 19.1 \\
\bottomrule
\end{tabular*}
\end{minipage}
\end{table}

\FloatBarrier
\vspace{-6pt}

\noindent Task-aware gating further omits coordinates whose reconstruction
errors have little downstream impact. Relative to quantization alone, these two mechanisms reduce the
storage rate by $30.6\%$--$67.6\%$ across the three benchmarks, with changes
in answer accuracy of at most $0.31$ percentage points. Thus, MemoWM's
storage gains do not come from quantization alone: they arise from using the
world model both to \textbf{code predictable information more efficiently and
to identify information that need not be stored at all}.

\looseness=-1
\noindent\textbf{Integration with M2A.}
We retrain the complete MemoWM storage pipeline---its Q-Former, quantizer,
and world model---for M2A, then replace raw evidence in M2A's
\mbox{RawMessageStore} with its encoded states. Across the three benchmarks,
this substantially reduces experience-specific storage while leaving answer
accuracy nearly unchanged. The small numerical gains on WMA and LME-V2 may
reflect a cleaner representation: Q-Former compression can filter irrelevant
content, and fewer soft tokens may ease attention for the answer model.
These results expose a composable principle behind MemoWM:
\textbf{memory organization and memory representation need not be redesigned
together}. An existing system can preserve how it organizes, retrieves, and
reasons over experience, while a shared predictive prior determines how much
of each experience must be stored explicitly.

\FloatBarrier

\subsection{The Storage Trade-off Between World Models and Experience}
\label{sec:model-memory-tradeoff}

Persistent information in MemoWM is split between reusable world-model
capacity and recurring experience-specific storage. A larger world model
incurs a higher one-time parameter cost but can reduce the bits required for
each retained experience. We study this trade-off on AMA-Bench by varying
model capacity and comparing experience-specific rate with amortized total
storage.

\begin{figure}[!ht]
\centering
\begin{minipage}[t]{0.49\textwidth}
\centering
\includegraphics[width=\linewidth,trim=0 0 0 2bp,clip]{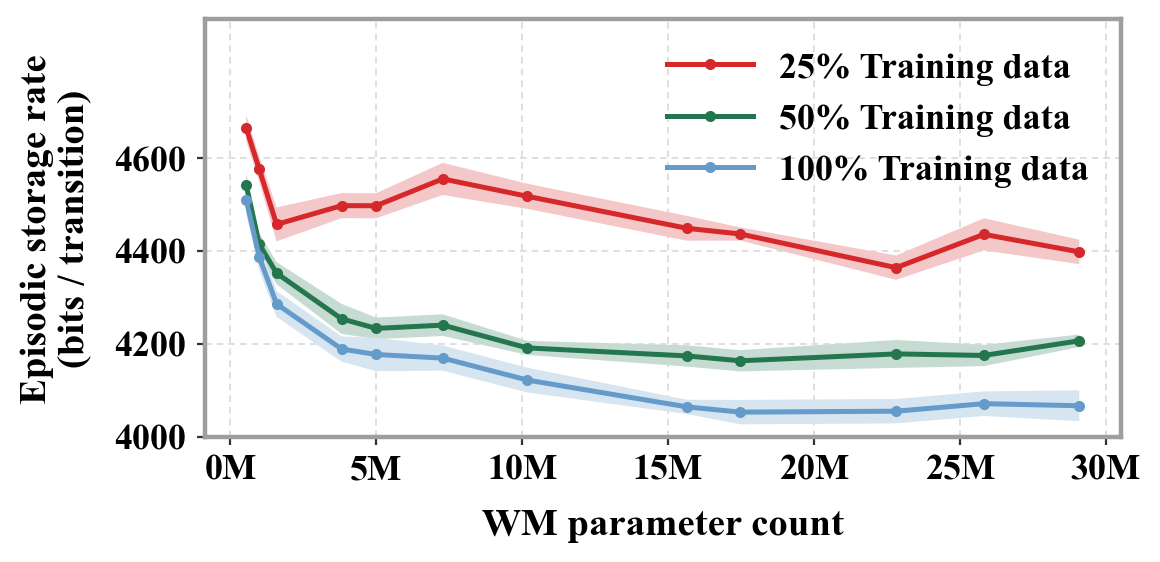}
\end{minipage}\hfill
\begin{minipage}[t]{0.49\textwidth}
\centering
\includegraphics[width=\linewidth,trim=0 0 0 2bp,clip]{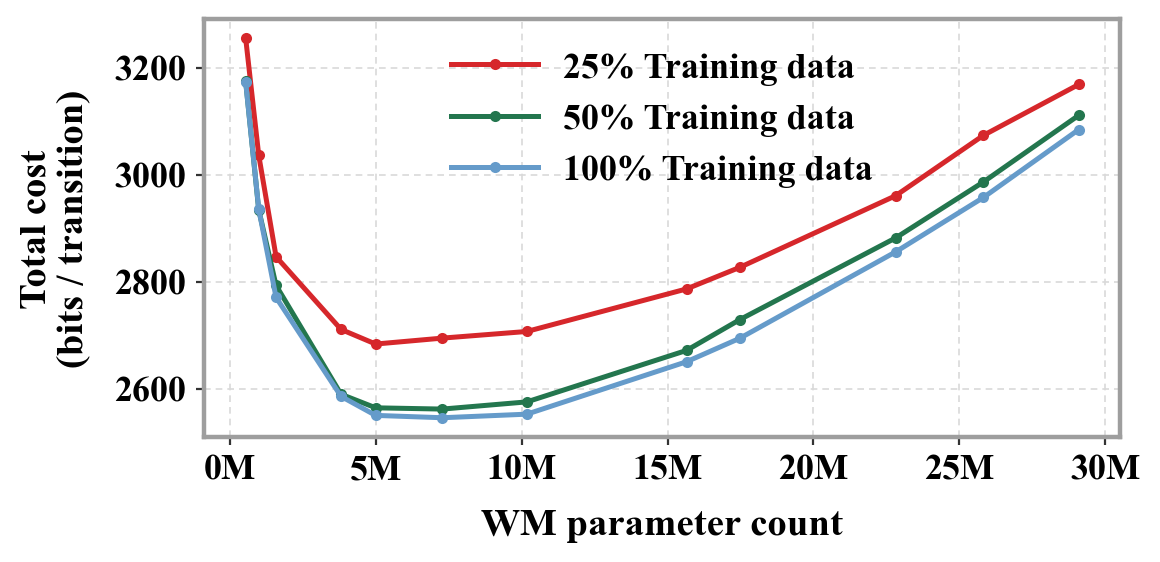}
\end{minipage}

\vspace{0.05em}
\begin{minipage}[t]{0.495\textwidth}
\centering
\includegraphics[width=\linewidth,height=0.47\linewidth]{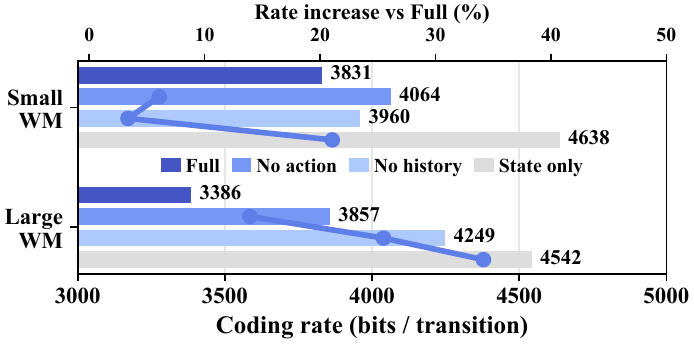}
\end{minipage}\hspace{-0.01\textwidth}
\begin{minipage}[t]{0.495\textwidth}
\centering
\includegraphics[width=\linewidth,height=0.47\linewidth]{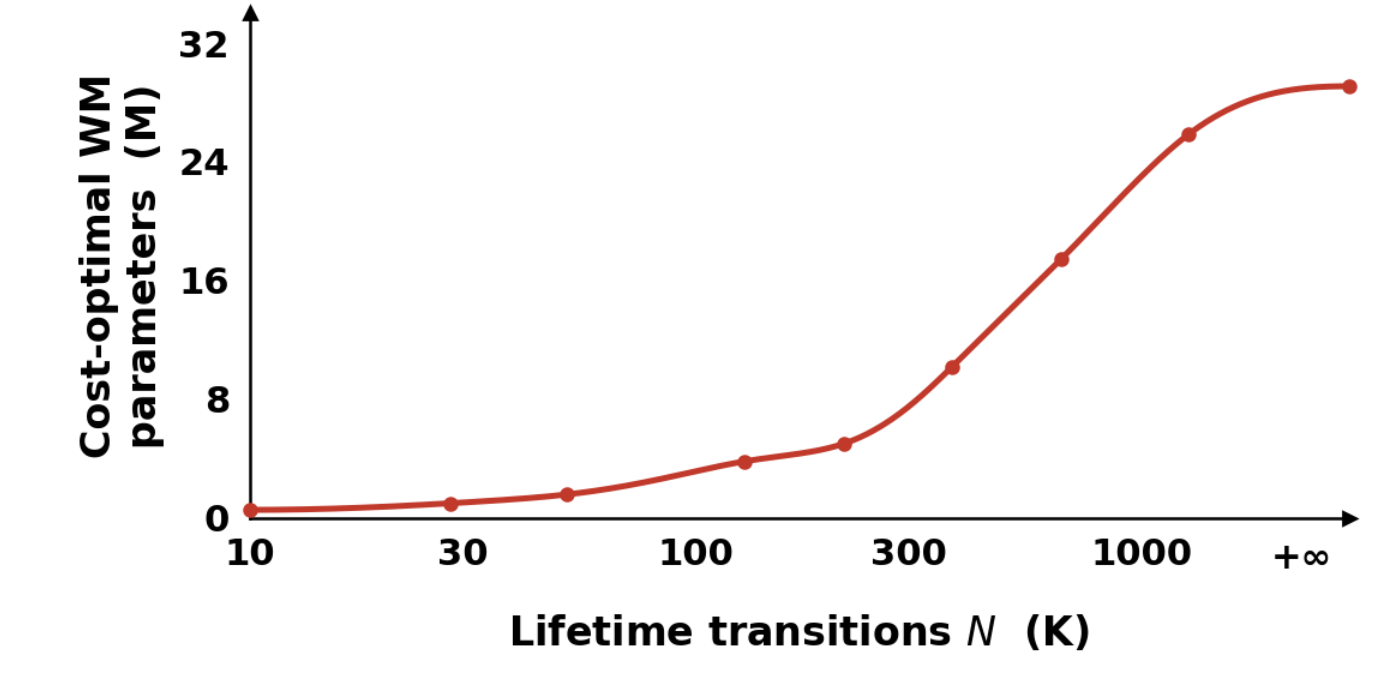}
\end{minipage}
\vspace{-0.4em}
\caption{(a) Upper left: all-coordinate
coding rate and (b) upper right: gated total rate with amortized FP16 weights,
both versus model size at three training-data fractions. (c) Lower left:
all-coordinate rate under four predictive contexts; blue curves show increases
over full context (upper axis). (d) Lower right: storage-optimal gated model
size versus retained transitions.}
\label{fig:model-memory-tradeoff}
\end{figure}

\noindent\textbf{Greater model capacity reduces recurring storage.}
Across all three training-data settings, larger world models generally lower
experience-specific storage. With full training data, the coding rate drops
from about $4{,}510$ to $4{,}070$ bits/transition. Because all coordinates are
retained, this reduction comes purely from more efficient conditional coding.
Larger world models extract more information from actions and history, as
removing either increases the rate far more at larger capacities.

\noindent\textbf{Total storage therefore has a finite optimal world-model
capacity.}
For $N$ retained transitions, the amortized cost is
$\bar S(P,N)=B_{\mathrm{shared}}(\Phi_P)/N+\bar R_{\mathrm{epi}}(P)$, where
$B_{\mathrm{shared}}(\Phi_P)=16P$ bits for FP16 weights. Larger models
reduce recurring episodic storage but incur greater one-time parameter cost,
yielding diminishing returns. Accordingly, the total cost first decreases and
then increases with capacity, with the evaluated optimum around $5$--$7$M
parameters (Figure~\ref{fig:model-memory-tradeoff}b).

\begin{figure}[!t]
\centering
\begin{minipage}[t]{0.494\textwidth}
\centering
\includegraphics[width=\linewidth]{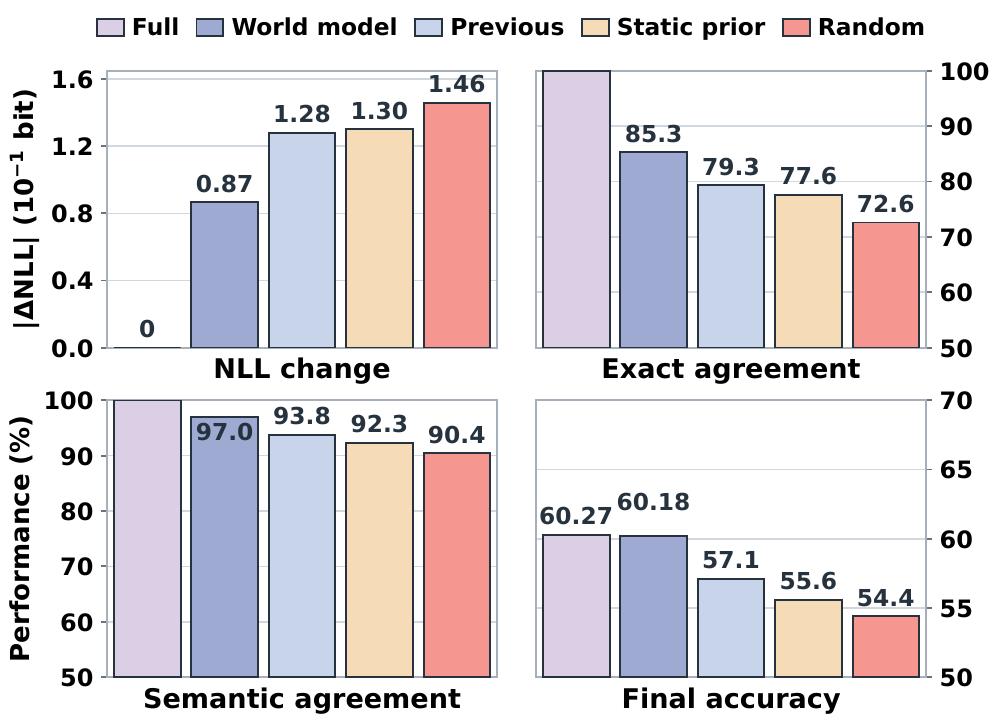}
\end{minipage}\hfill%
\begin{minipage}[t]{0.494\textwidth}
\centering
\includegraphics[width=\linewidth]{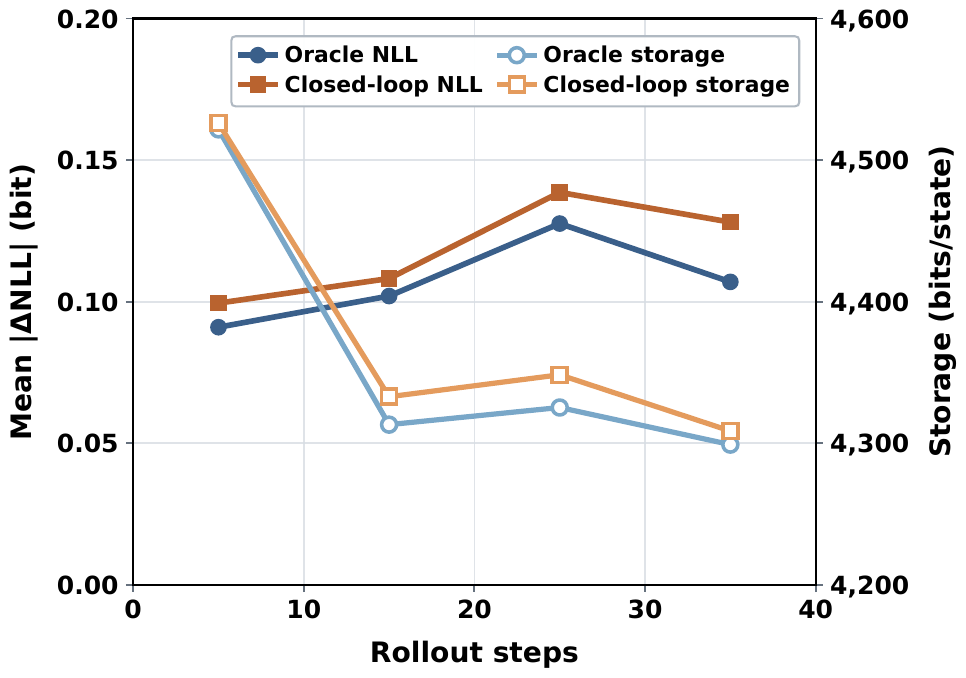}
\end{minipage}
\caption{Left: where decoder-side reconstruction quality comes from, with the
omitted coordinates and storage cost held fixed so that only the fill source
changes. Right: answer distortion and storage over four rollout-step ranges for
teacher-forced oracle and closed-loop reconstruction.}
\label{fig:wm-source-ablation}
\label{fig:closed-loop-depth}
\end{figure}

\noindent\textbf{Longer operating lifetimes favor larger world models.}
As more experience is retained, the same world model is amortized over more
transitions. Holding the experience-rate curve fixed, increasing $N$ reduces
$B_{\mathrm{shared}}(\Phi_P)/N$, allowing the recurring savings of larger
models to outweigh their additional parameter cost. Accordingly, the
storage-optimal capacity increases with memory horizon
(Figure~\ref{fig:model-memory-tradeoff}d), from approximately $0.54$M
parameters at $10$K retained transitions toward $29$M at long horizons.
Thus, longer-lived agents can justify greater investment in reusable world
knowledge because its cost is paid once while its storage savings recur
across experience.

\subsection{Task-Aware Risk Improves the Rate--Utility Trade-off}
\label{sec:risk-tradeoff}

We test whether the factorized risk $e_{t,j}V_j$ improves storage allocation
against four gating baselines: Static-$e$ (state-independent error estimate),
Uniform-$V$ (coordinate-independent task impact), Random, and Uncertainty
(entropy only). We tune $\lambda$ separately for MemoWM and each gating
baseline to match $|\Delta\mathrm{NLL}|=1.536$ in closed-loop reconstruction,
with All-send as the zero-distortion reference.
On held-out trajectories, the task-impact estimate $V_j$ can be shared across
states, and entropy orders state-dependent prediction risk, supporting both
risk components as reusable signals. Damage also remains low relative to
random omission under independently constructed unseen-query variants,
indicating that the shared task-impact prior remains stable for rare query
formulations. The detailed numerical analysis and
theoretical development are provided in
\hyperref[app:shared-task-prior]{Appendix~\ref*{app:shared-task-prior}}, and
estimator implementation details are given in
\hyperref[app:method-implementation]{Appendix~\ref*{app:method-implementation}}.

\begin{table}[!ht]
\noindent
\begin{minipage}[t]{0.47\textwidth}
\vspace{0pt}
\raggedright
\looseness=-1
\textbf{Joint risk allocation.} Under matched distortion, MemoWM has the
lowest rate (Table~\ref{tab:risk-tradeoff}). Static-$e$ and Uniform-$V$ incur
$17.0\%$ and $25.5\%$ overhead, respectively, showing that both state-dependent
prediction risk and coordinate-specific task impact improve allocation.
The larger overheads of Random and Uncertainty ($33.1\%$ and $42.0\%$) further
support combining prediction risk with task impact to decide which coordinates
to store.
\end{minipage}\hfill
\begin{minipage}[t]{0.51\textwidth}
\vspace{-\abovecaptionskip}
\caption{Task-aware gating at matched distortion. Rate: bits/state;
Rate $\uparrow$: increase over MemoWM.}
\label{tab:risk-tradeoff}
\centering
\footnotesize
\setlength{\tabcolsep}{3.5pt}
\begin{tabular*}{\linewidth}{@{\extracolsep{\fill}}lrrr@{}}
\hline
Gate & $|\Delta\mathrm{NLL}|$ & Rate & Rate $\uparrow$ \\
\hline
All-send & 0.000 & 3737.5 & $+58.8\%$ \\
\textbf{MemoWM} & 1.536 & \textbf{2354.2} & -- \\
Static-$e$ & 1.536 & 2754.3 & $+17.0\%$ \\
Uniform-$V$ & 1.536 & 2954.8 & $+25.5\%$ \\
Random & 1.536 & 3134.2 & $+33.1\%$ \\
Uncertainty & 1.536 & 3342.1 & $+42.0\%$ \\
\hline
\end{tabular*}
\end{minipage}
\end{table}

\FloatBarrier

\subsection{Reconstruction Quality and Closed-Loop Stability}
\label{sec:reconstruction-stability}

\noindent\textbf{World-model reconstruction preserves task-relevant
information.}
Holding the omitted coordinates and storage budget fixed, world-model filling
yields the best compressed reconstruction among the tested alternatives
(Figure~\ref{fig:wm-source-ablation}, left). Compared with previous-state
copying, it reduces answer-NLL distortion from $0.1280$ to $0.0867$ bit and
improves final accuracy from $57.10\%$ to $60.18\%$, only $0.09$ points below
full quantized memory. Despite exact recovery of only $14.2\%$ of omitted
codes, semantic agreement reaches $97.0\%$, showing that task-level fidelity
does not require coordinate-level reconstruction.

\noindent\textbf{Closed-loop reconstruction remains stable.}
Because reconstructed states are recursively fed back into the world model, we
compare closed-loop prediction with a teacher-forced oracle using true history.
Over rollouts up to $40$ steps, closed-loop reconstruction adds only
$0.006$--$0.021$ bit of answer-NLL distortion and $4.4$--$23.1$ bits/state of
storage, indicating limited error accumulation over the evaluated horizon
(Figure~\ref{fig:closed-loop-depth}, right).
\FloatBarrier

\section{Conclusion}
\label{sec:conclusion}

We view long-term memory as a division of labor between shared world knowledge
and episodic experience: what the world model can reliably recover need not be
stored repeatedly. Based on this principle, MemoWM uses world-model predictions
as a shared prior and stores only information that is both hard to recover and
important for downstream tasks. Experiments show that this substantially
reduces recurring memory cost while preserving task performance, and that the
optimal balance between world-model capacity and episodic storage depends on
the agent's memory horizon.

\subsection*{AI use statement}

We used generative AI tools for English translation, language editing,
text shortening, LaTeX formatting, and manuscript consistency checks.
Separately, language models were used within the research pipeline for
training-data collection and supervision, answering, and evaluation, as
described in the paper. The authors checked all citations, equations,
experimental settings, and results, reviewed all AI-assisted text, and
made the final scientific judgments. The authors take full responsibility
for the paper.

\subsection*{Ethics statement}

This work studies compressed memory for long-term agents using benchmark
evaluations, trajectories collected in self-hosted WebArena environments,
and synthetic training data. The MemEye training data use newly generated
virtual patients and screens rather than reusing evaluation patient records.
The health-related experiments assess memory and visual reasoning and do
not establish clinical validity. Persistent agent memory can retain
sensitive information, and predictive reconstruction can introduce
incorrect details or omit important evidence. Real-world deployment should
therefore include access controls, appropriate consent and deletion
mechanisms, and verification of consequential recalled information.
Compression alone does not provide a privacy guarantee. We describe
training--evaluation overlap checks and the limitations of the synthetic
MemEye data in Appendix~\ref{app:webarena-data}.

\subsection*{Reproducibility statement}

Section~\ref{memowm} specifies the retention rule, shared-history
reconstruction, and calibration procedure. Section~\ref{sec:experiments}
and Appendix~\ref{appendix-detailed-methods-and-evaluation-protocol-for-memowm}
describe the experimental setup, model architecture, discretization, and
offline fitting. Appendix~\ref{app:webarena-data} describes training-data
construction and overlap checks; Appendices~\ref{app:storage-accounting},
\ref{app:entropy-coding}, and~\ref{app:evaluation-settings} document storage
accounting, a physical entropy-coding validation, and evaluation settings,
respectively. Main-table answer accuracies are reported as means and
standard deviations over four random seeds. The code repository
is linked at the beginning of the paper.

\bibliography{iclr2027_conference}
\bibliographystyle{iclr2027_conference}

\clearpage
\appendix
\section{Detailed Methods and Evaluation Protocol for
MemoWM}\label{appendix-detailed-methods-and-evaluation-protocol-for-memowm}

This appendix supplements Section \ref{memowm} with implementation details, derives
the trade-off between model capacity and episodic storage, analyzes
state-dependent gating, and specifies the data construction, storage
accounting, and evaluation protocols.

Throughout, \(z_t=E(o_t)\) denotes the complete discrete reference state
and \(\tilde z_t\) the reconstructed state after retention decisions.
Answer-loss comparisons against complete quantized memory measure the
additional effect of predictive omission; they do not include
information lost when the original observation is first quantized.

\subsection{State Representation and
Discretization}\label{a.1-state-representation-and-discretization}

Let \(F_\rho\) be the continuous observation encoder. It reads frozen
Qwen3.5-9B layer-16 features. The Q-Former uses 32 learned queries, four
cross-attention layers, eight heads, and hidden width 1,024. It outputs
a \(32\times512\) state representation, uses query-key normalization,
and disables query self-attention. The soft-token connector is trained
with the Q-Former and frozen before counterfactual task-impact
measurements.

The Q-Former is trained with two levels of readout supervision.
Overall-observation probes ask about visible content, page layout, input
values, and control states. Detail questions and answers are generated
by an LLM from WebArena screenshots and target local facts visible in
the current observation; they are not copied from benchmark task
annotations. For a training question \(r\), a teacher that reads the
full observation produces \(p_T(\cdot\mid o,r)\), while a latent student
produces \(p_S(\cdot\mid F_\rho(o),r)\) through the trainable connector.
Overall probes use distillation from teacher-generated continuations.
Detail questions use both cross-entropy with the reference answer and
teacher-distribution distillation. Other observations from the same
session are used as negatives in a symmetric InfoNCE objective:

\[
\mathcal L_{\rm QF}=0.5\mathcal L_{\rm KL,obs}
+\mathcal L_{\rm CE}
+0.3\mathcal L_{\rm KL,q}
+\mathcal L_{\rm sem}.
\]

The InfoNCE temperature is 0.05. Questions provide readout supervision
only and are not inputs to the encoder; memory formation always computes
the query-independent \(F_\rho(o)\). Three groups of overall-observation
probes are used for gradient training, and a fourth group is reserved
for checkpoint selection and observation-fidelity evaluation.

Let \(Q_\omega\) be the frozen quantizer, so that
\(E=Q_\omega\circ F_\rho\) and \(\tilde x_t=D_\omega(\tilde z_t)\). OPQ
first standardizes each slot and channel, then applies a shared
\(512\times512\) orthogonal rotation. Each slot is divided into 32
subspaces of dimension 16, with 64 codewords per subspace. This gives
1,024 discrete coordinates and a fixed-width reference of 6,144 bits per
observation. The reported codebook uses 520,110 training states, 25
k-means iterations, and seed 0. The world model and task-impact labels
are generated in this frozen OPQ coordinate system. The difference
between \(D_\omega(z_t)\) and \(D_\omega(\tilde z_t)\) therefore comes
only from coordinates replaced by world-model predictions.

\subsection{World Model and Shared-History
Reconstruction}\label{a.2-world-model-and-shared-history-reconstruction}

The world model reads at most 16 historical states. Each subspace code
has an 8-dimensional embedding; the 32 codes in one slot are
concatenated into a 256-dimensional state token and combined with
action, relative-time, slot-identity, and validity embeddings. The
backbone is a 12-layer Transformer with eight heads, width 256, MLP
width 4,096, and dropout 0.1. Attention is causal across time blocks but
permits the 32 slots within one block to interact. The final valid block
predicts all 1,024 coordinates of the next state in parallel.

Structured action encoding contains 17 action types, 64 bins for each
absolute position coordinate plus a missing-value category, a signed-log
representation of relative displacement, and a GRU encoding of up to 40
payload bytes. The action is aligned with the transition from the last
historical state to the state being predicted. The world model has
29,092,096 parameters. The reported training uses 495,527 transitions
for 60,000 steps, with true codes in the history. The objective combines
next-state code negative log-likelihood with a slot-validity term. All
32 slots are valid in the selected interface, so all 1,024 codes
contribute to prediction and code-length computation. Training uses
dropout 0.1; dropout is disabled during writing and reconstruction to
keep inference deterministic.

During training the model uses true-code histories; during writing and
decoding it uses \(h_t=(\tilde z_{<t},u_{<t})\). In both cases, all
distributions for the current state are computed before any current true
coordinate is read. The factorized coding model is

\[
p_\theta(z_t\mid h_t)=\prod_{j=1}^{J}p_{t,j}(z_{t,j}).
\]

This factorization is a coding choice and does not claim that the
coordinates are independent in the true observation distribution. For
\(t\ge1\), the output copy gate mixes the previous reconstructed code
with the new categorical prediction:

\[
p_{t,j}(c)=\gamma_{t,j}\mathbf 1\{c=\tilde z_{t-1,j}\}
+(1-\gamma_{t,j})\operatorname{Softmax}(\ell_{t,j})_c,
\qquad 0<\gamma_{t,j}<1.
\]

The gate makes persistent interface content more likely to receive a
short code. With finite logits, every category has positive probability,
so the ideal code length is defined. The symbol \(\pi\) is reserved for
the retention rule.

At \(t=0\), the initial state is stored in full and both sides set
\(\tilde z_0=z_0\). For \(t\ge1\), the action \(u_{t-1}\) is available
before state \(z_t\) is recovered. Both writer and decoder compute the
same distributions, modal predictions, entropies, and retention mask
from the common reconstructed history. The reconstructed state is

\[
\tilde z_{t,j}=
\begin{cases}
z_{t,j},&m_{t,j}=1,\\
\hat z_{t,j},&m_{t,j}=0.
\end{cases}
\]

The writer feeds back \(\tilde z_t\), rather than the original \(z_t\),
before processing the next step. This makes later predictions available
to the decoder. If the initial state uses \(\lceil\log_2C\rceil\) bits
per coordinate, the episodic storage is

\[
R_{\mathrm{epi}}^\pi(\tau)
=J\lceil\log_2C\rceil
+\sum_{t=1}^{T}\sum_{j=1}^{J}m_{t,j}r_{t,j}
+B_{\mathrm{aux}}^\pi(\tau).
\]

Here \(B_{\mathrm{aux}}^\pi\) denotes per-episode auxiliary content.
Under the accounting protocol in this appendix, it includes actions,
optional symbol anchors, and other fields needed for decoding that
cannot be recovered losslessly from counted content; regenerable
retrieval vectors and indices are excluded from the main table but may
be reported as separate implementation overhead. Shared model storage is
counted separately. Retention masks need not be transmitted because the
decoder recomputes them from its own reconstructed history.

\subsection{Risk Estimation, Retrieval Training, and Threshold
Calibration}\label{app:method-implementation}

\subsubsection{Risk estimation and
calibration}\label{a.3.1-risk-estimation-and-calibration}

Counterfactual labels use the same frozen Reader Bridge and Reader for a
complete OPQ state and a state with one coordinate replaced by its modal
prediction. For every reference-answer token, teacher-forced NLL is
divided by \(\ln 2\) and reported in bits. The two branches use
identical questions and answer tokens, and the implementation records
the measurements in float32.

For coordinate \(j\), let \(\mathcal I_j\) be its state-question
annotation set. Define \(A_{i,j}=\mathbf1\{\hat z_{i,j}\ne z_{i,j}\}\)
as the prediction-error indicator and let \(\Delta_{i,j}\) be the
bit-valued answer-loss change caused by replacing that coordinate with
its prediction. Let

\[
n_j^{\rm err}=\sum_{i\in\mathcal I_j}A_{i,j}.
\]

When \(n_j^{\rm err}>0\), the conditional task impact is

\[
\widehat V_j
=\frac{\sum_{i\in\mathcal I_j}A_{i,j}\Delta_{i,j}}
{n_j^{\rm err}}.
\]

Only incorrect predictions enter this average. An unconditional average
already contains the offline error frequency, so multiplying it by
another error probability would count that factor twice.

The deployed error lookup bins predictive entropy at 0.25 bits. For a
nonempty bin, calibration records from all states and coordinates are
pooled and the modal prediction-error frequency is used. For an empty
bin, the lookup falls back to the marginal calibration error rate of
coordinate \(j\). Thus \(g_j(H)\) uses a shared value for populated bins
rather than fitting a separate curve for every coordinate. The frozen
lookup and empty-bin rule are shared by writer and decoder. Online risk
is

\[
\hat d_{t,j}=g_j(H_{t,j})\widehat V_j.
\]

\subsubsection{Retrieval representation
learning}\label{a.3.2-retrieval-representation-learning}

For one observation, the complete and reconstructed discrete states are
mapped through \(D_\omega\) and then through a retrieval head
\(K_\eta\). The head applies per-slot LayerNorm, learned attention
pooling, a \(512\rightarrow1024\rightarrow4096\) MLP, and \(\ell_2\)
normalization. The state encoder, OPQ, world model, and task-impact
statistics are frozen, while the retrieval head is trained from random
initialization. A retrieval head that may have been jointly trained in a
Q-Former checkpoint is not reused.

Let \(z_i^+\) denote the complete quantized state for observation \(i\),
and let \(\hat z_i\) denote its coordinate-wise world-model prediction.
The complete and reconstructed views are paired with the same frozen
Qwen3-VL full-observation teacher embedding \(t_i\in\mathbb R^{4096}\).
The teacher is used only for offline supervision and is not stored as an
online memory key. The paired inputs are \(x_i^{\rm f}=D_\omega(z_i^+)\)
and \(x_i^{\rm g}=D_\omega(m_i\odot z_i^++(1-m_i)\odot\hat z_i)\), where
\(m_i\) is the retention mask. We write the normalized retrieval-head
output as \(h(x)\). For a batch, let \(S_{ab}=h(x_a)^\top t_b/0.05\).

\[
\mathcal L_{\rm ret}(x,t)=
\tfrac12\left[\operatorname{CE}(S,I_B)+\operatorname{CE}(S^\top,I_B)\right]
+\frac1B\sum_i\left[1-\cos(h(x_i),t_i)\right].
\]

The paired-view objective is

\[
\mathcal L_{\rm head}=
\tfrac12\mathcal L_{\rm ret}(x^{\rm f},t)
+\tfrac12\mathcal L_{\rm ret}(x^{\rm g},t)
+\frac{0.1}{B}\sum_i
\left[1-\cos\!\left(h(x_i^{\rm g}),
\operatorname{sg}(h(x_i^{\rm f}))\right)\right].
\]

The stop-gradient operator prevents the consistency term from updating
the complete-view target; the complete view is still trained by its own
retrieval loss. The teacher-alignment and in-batch negative terms
encourage observation-discriminative keys and reduce the risk of
representation collapse; they are not a formal guarantee against
collapse. AdamW uses batch size 64, learning rate \(3\times10^{-4}\),
weight decay \(10^{-4}\), and gradient-norm cap 5. The checkpoint is
selected by validation dual-view loss. In WorldMemArena,
Qwen3-VL-Embedding-8B encodes queries, and the top 10 memories within
the same session are selected by cosine similarity and passed to the
Reader Bridge.

An optional exact-symbol anchor extracts element ID, role, and visible
name from accessibility-tree paths. A lightweight OCR path extracts
dates, URLs, and typed numbers from screenshots. These symbols do not
enter OPQ coordinates or retrieval keys; they are converted to Reader
token embeddings only when their corresponding observation is retrieved.
Their storage cost is counted only when this path is enabled.

\subsubsection{Reconstructed-memory
reading}\label{a.3.3-reconstructed-memory-reading}

For future question \(q\), complete memory \(Z\) and reconstructed
memory \(\widetilde Z^\pi\) use the same retrieval head, ranking rule,
Reader Bridge, Reader, and answer-token scoring convention. Their
retrieved states may differ, so the full-memory distortion includes both
retrieval changes and changes in the evidence read by the Reader. An
optional symbol anchor enters the Reader only with its corresponding
retrieved observation.

\subsubsection{Threshold
calibration}\label{a.3.4-threshold-calibration}

For each candidate \(\lambda\in\Lambda\), calibration trajectories are
reconstructed from their complete initial states using the risk rule and
shared-history recursion. Fixed retrieval and reading components then
measure the answer-loss change relative to complete quantized memory and
the episodic storage rate. The candidate with minimum measured storage
among those satisfying the average distortion constraint is selected and
frozen for held-out evaluation.

Candidates are evaluated with separate closed-loop reconstructions
because early omissions can change later histories, code lengths, and
masks. On the same calibration episodes, the observation-normalized
denominator is unchanged across candidates, so minimizing rate is
equivalent to minimizing total episodic storage. At \(\lambda=0\), the
tie rule retains every coordinate; reconstructed memory equals the
complete quantized reference and the additional answer loss is zero.

\subsubsection{Formal analysis of the retention
rule}\label{a.3.5-formal-analysis-of-the-retention-rule}

For an integrable local omission loss, a correct prediction leaves the
discrete state unchanged, so \(\Delta_{t,j}=0\) whenever \(A_{t,j}=0\).
Conditional total expectation therefore gives

\[
\begin{aligned}
\mathbb E_{\rm obs}[\Delta_{t,j}\mid h_t]
={}&\Pr_{\rm obs}(A_{t,j}=1\mid h_t)\cdot\mathbb E_{\rm obs}[\Delta_{t,j}\mid A_{t,j}=1,h_t].
\end{aligned}
\]

If the conditional error probability is zero, the product is defined as
zero. This identity does not assume independence between error
occurrence and error impact. Approximation enters only when the two
context-dependent factors are replaced by \(g_j(H_{t,j})\) and
\(\widehat V_j\).

For a fixed history, the difference between retaining and omitting
coordinate \(j\) in the proxy objective is

\[
\widehat{\mathcal J}_{t,j}(1)-\widehat{\mathcal J}_{t,j}(0)
=\lambda H_{t,j}-\widehat d_{t,j}.
\]

Retaining the true value when this difference is nonpositive, including
ties, minimizes the coordinate-level proxy for that fixed history. Since
the proxy is additive across coordinates, the coordinate decisions
jointly minimize the fixed-history proxy. This does not establish global
optimality for a complete episode: simultaneous omissions may alter
retrieval and subsequent predictive histories, which is why threshold
calibration uses complete recurrent reconstruction.

The actual code length \(r_{t,j}=-\log_2p_{t,j}(z_{t,j})\) depends on
the true value known to the writer but unavailable to the decoder before
the value is read. Predictive entropy \(H_{t,j}\) depends only on the
shared predictive distribution and can therefore be used in a shared
decision. If \(\mu_{t,j}(\cdot\mid h_t)\) is the true conditional
distribution,

\[
\mathbb E_{\mu_{t,j}}[r_{t,j}\mid h_t]
=H(\mu_{t,j})+D_{\mathrm{KL},2}(\mu_{t,j}\Vert p_{t,j}),
\]

which is generally different from \(H_{t,j}=H(p_{t,j})\). Entropy is
therefore a jointly computable proxy for decision-time cost; episodic
storage is still charged using the ideal lengths of retained true
values.

Finally, assume identical initial states, actions, frozen models and
lookup tables, coordinate ordering, numerical computation, tie rules,
and exact decoding of retained values. The histories agree at \(t=0\).
If reconstructed histories agree through \(t-1\), both sides compute the
same distributions, entropies, masks, and modal predictions at \(t\).
They use the same decoded value at retained coordinates and the same
prediction at omitted coordinates, so \(\tilde z_t\) also agrees.
Induction establishes synchronization of the reconstructed histories; it
does not establish equality between reconstructed and original states.

\subsection{Model Capacity and Episodic-Storage
Trade-off}\label{app:model-memory-tradeoff}

\subsubsection{Total storage and cost per
transition}\label{a.4.1-total-storage-and-cost-per-transition}

Let \(\Phi_P\) be a world model with \(P\) parameters and let \(\pi_P\)
be its associated retention rule. For an episode \(\tau\),

\[
\begin{aligned}
R_{\mathrm{epi}}^{\pi_P}(\tau)
={}&J\lceil\log_2 C\rceil
+\sum_{t=1}^{T}\sum_{j=1}^{J}m_{t,j}^{\pi_P}r_{t,j}^{\Phi_P}+B_{\mathrm{aux}}^{\pi_P}(\tau).
\end{aligned}
\]

Let \(B_{\mathrm{shared}}(\Phi_P)\) be the shared-model storage cost.
For a collection \(\mathcal T=\{\tau_i\}_{i=1}^{K}\) with
\(N=\sum_iT_i>0\) transitions,

\[
S(P;\mathcal T)=B_{\mathrm{shared}}(\Phi_P)
+\sum_{i=1}^{K}R_{\mathrm{epi}}^{\pi_P}(\tau_i),
\]

and the episodic rate per transition is

\[
\bar R_{\mathrm{epi}}(P;\mathcal T)
=\frac{\sum_iR_{\mathrm{epi}}^{\pi_P}(\tau_i)}
{\sum_iT_i}.
\]

This subsection uses transitions as the denominator. If the same
collection is normalized by observations, the denominator is \(N+K\)
because each of the \(K\) episodes contributes an initial observation.
The storage table uses the observation-based denominator specified in
Section A.7.

The corresponding total cost per transition is

\[
\bar S(P,N)=\frac{B_{\mathrm{shared}}(\Phi_P)}{N}
+\bar R_{\mathrm{epi}}(P).
\]

Assume the shared cost is continuous, strictly increasing, and
unbounded, with \(\dot B(P)>0\) and \(\ddot B(P)\ge0\). The experiments
use FP16 weight storage, so \(B_{\mathrm{shared}}(\Phi_P)=16P\). For the
horizon analysis, the measured episodic-rate curve and episode-length
distribution are held fixed; only the number \(N\) of retained
transitions changes. With a fixed state interface, every episode must
retain its initial state, giving the rate floor

\[
\bar R_{\mathrm{epi}}(P)\ge
R_{\mathrm{floor}}
\equiv\frac{KJ\lceil\log_2C\rceil}{N}.
\]

\subsubsection{Diminishing returns and a finite
optimum}\label{a.4.2-diminishing-returns-and-a-finite-optimum}

If \(\bar R_{\mathrm{epi}}'(P)<0\) and \(\bar R_{\mathrm{epi}}''(P)>0\),
capacity has diminishing returns. Writing
\(g(P)=-\bar R_{\mathrm{epi}}'(P)\),

\[
\int_{P_{\min}}^{P}g(u)\,du
=\bar R_{\mathrm{epi}}(P_{\min})-\bar R_{\mathrm{epi}}(P)
\le\bar R_{\mathrm{epi}}(P_{\min})-R_{\mathrm{floor}},
\]

so \(g(P)\to0\). For fixed \(N\),

\[
\frac{\partial\bar S}{\partial P}
=\frac{\dot B(P)}{N}-g(P),
\qquad
\frac{\partial^2\bar S}{\partial P^2}
=\frac{\ddot B(P)}{N}+\bar R_{\mathrm{epi}}''(P)>0.
\]

An interior optimum therefore satisfies

\[
-\bar R_{\mathrm{epi}}'(P^*)=\frac{\dot B(P^*)}{N}.
\]

If this crossing does not occur after \(P_{\min}\), the smallest
candidate is optimal.

\subsubsection{Break-even between two
models}\label{a.4.3-break-even-between-two-models}

For two capacities \(P_b>P_a\), define
\(\Delta B_{\mathrm{shared}}=B_{\mathrm{shared}}(\Phi_{P_b})-B_{\mathrm{shared}}(\Phi_{P_a})>0\)
and
\(\Delta\bar R_{\mathrm{epi}}=\bar R_{\mathrm{epi}}(P_a)-\bar R_{\mathrm{epi}}(P_b)\).
When \(\Delta\bar R_{\mathrm{epi}}>0\), the larger model becomes cheaper
exactly when

\[
N>
\frac{\Delta B_{\mathrm{shared}}}{\Delta\bar R_{\mathrm{epi}}}.
\]

If \(\Delta\bar R_{\mathrm{epi}}\leq0\), the smaller model is cheaper
for every positive retention scale.

\subsubsection{Deployment horizon and preferred
capacity}\label{a.4.4-deployment-horizon-and-preferred-capacity}

Under the stated differentiability, convexity, and fixed-rate
assumptions, the interior preferred capacity increases with the
deployment horizon. In the smooth case,

\[
\frac{dP^*}{dN}
=\frac{\dot B(P^*)}
{N^2\bar R_{\mathrm{epi}}''(P^*)+N\ddot B(P^*)}>0.
\]

For a discrete candidate set, pairwise optimality comparisons give the
same monotonic result: if \(N_2>N_1\), then an optimal capacity at
\(N_2\) is no smaller than an optimal capacity at \(N_1\). After a
consistent tie-breaking rule is fixed, the discrete optimum grows in
steps.

\subsubsection{Implications for long-term
memory}\label{a.4.5-implications-for-long-term-memory}

Applying the cost calculation to the measured curves gives a local
optimum near 5--7M parameters and a shift from roughly 0.54M parameters
at short horizons toward about 29M at long horizons; these values are
empirical results for the reported candidate set and deployment
distribution.

\subsection{State-Dependent Gating and a Shared Task-Impact
Prior}\label{app:shared-task-prior}

\subsubsection{Common coordinate space and conditional
damage}\label{a.5.1-common-coordinate-space-and-conditional-damage}

The task-impact estimate \(\widehat V_j\) is shared across states, while
\(g_j(H_{t,j})\) changes with the current predictive distribution. The
frozen Q-Former and OPQ provide a consistent ordering of slots and
subspaces; a shared coordinate need not represent the same semantic
concept in every observation. Let the transformed representation contain
\(S\) slots of width \(d\), each divided into \(M\) subspaces, so the
number of discrete coordinates is \(J=SM\). Here \(S=32\), \(d=512\),
\(M=32\), and \(J=1024\); each subspace has dimension 16. After
flattening the representation,

\[
\mathbb R^{Sd}=\bigoplus_{j=1}^{J}\mathcal V_j.
\]

Here \(\mathcal V_j\) is the 16-dimensional subspace of coordinate
\(j\), and \(P_j\) below denotes the orthogonal projection onto that
subspace. Let \(r_i^+\) be the transformed representation reconstructed
from the complete quantized state. Replacing only coordinate \(j\) gives
\(\delta_{i,j}\in\mathcal V_j\) and \(r_i^{(-j)}=r_i^++\delta_{i,j}\).
If \(F_i(r)\) denotes the natural-log loss of the frozen single-state
Reader after inverse transformation and connector application, then

\[
\Delta_{i,j}=
\frac{|F_i(r_i^++\delta_{i,j})-F_i(r_i^+)|}{\ln2}.
\]

For \(A_{i,j}=\mathbf1\{\hat z_{i,j}\ne z_{i,j}\}\), define
\(V_j^{\rm ref}=\mathbb E_{\rm obs}[\Delta_{i,j}\mid A_{i,j}=1]\) when
the conditioning event has positive probability. If the Reader loss has
an \(L\)-Lipschitz gradient along the replacement segment, then

\[
\Delta_{i,j}\leq\frac{1}{\ln2}
\left(|\langle P_j\nabla F_i(r_i^+),\delta_{i,j}\rangle|
+\frac L2\|\delta_{i,j}\|_2^2\right).
\]

Define

\[
S_j^{\rm err}=\mathbb E_{\rm obs}[\|P_j\nabla F_i(r_i^+)\|_2^2\mid A_{i,j}=1],
\qquad
A_j^{\rm err}=\mathbb E_{\rm obs}[\|\delta_{i,j}\|_2^2\mid A_{i,j}=1].
\]

Then

\[
V_j^{\rm ref}\leq\frac{1}{\ln2}
\left(\sqrt{S_j^{\rm err}A_j^{\rm err}}
+\frac L2 A_j^{\rm err}\right).
\]

Across a family of trajectory-conditioned distributions, define
\(V_j(Q)=\mathbb E_Q[\Delta_{i,j}\mid A_{i,j}=1]\) whenever the
conditioning event has positive probability. If, for every such \(Q\),
\(S_j^{\rm err}(Q)\le s_j^2\) and \(A_j^{\rm err}(Q)\le a_j^2\), set

\[
b_j=\frac{s_ja_j+(L/2)a_j^2}{\ln2},
\qquad
\bar b=\frac1J\sum_{j=1}^{J}b_j.
\]

Then \(V_j(Q)\le b_j\), and for any \(\varepsilon>\bar b\) the common
low-impact set \(\mathcal J_\varepsilon=\{j:b_j\le\varepsilon\}\)
satisfies

\[
|\mathcal J_\varepsilon|\ge
J\left(1-\frac{\bar b}{\varepsilon}\right)>0.
\]

The bound provides a sufficient condition for a shared set of
low-utility coordinates to remain low-impact across the specified family
of distributions. MemoWM estimates the combined effect directly with
counterfactual answer loss. Under the reported training--held-out split
and comparison size, selected coordinates have approximately
\(1/166\)--\(1/176\) of the error-conditional damage of same-size random
coordinates; this is an empirical result for that evaluation setting,
not a distribution-free guarantee.

\subsubsection{Entropy calibration and gating
stability}\label{a.5.2-entropy-calibration-and-gating-stability}

For entropy bin \(\mathcal B_b\), the pooled error estimate is

\[
\widehat e_b=
\frac{\sum_{(i,k):H_{i,k}\in\mathcal B_b}A_{i,k}}
{|\{(i,k):H_{i,k}\in\mathcal B_b\}|}.
\]

For a nonempty bin, the deployment value \(\widehat e_{j,b}\) uses this
pooled frequency; for an empty bin, it falls back to the marginal
calibration error rate of coordinate \(j\). Thus \(g_j(H)\) combines a
shared estimate for populated bins with a coordinate-specific fallback.
For bins with positive mass and positive conditional error probability,
define

\[
\begin{aligned}
U_{j,b}^{*}&=\mathbb E[\Delta_{t,j}\mid H_{t,j}\in\mathcal B_b],\\
e_{j,b}^{*}&=\Pr(A_{t,j}=1\mid H_{t,j}\in\mathcal B_b),\\
V_{j,b}^{*}&=\mathbb E[\Delta_{t,j}\mid A_{t,j}=1,H_{t,j}\in\mathcal B_b].
\end{aligned}
\]

Because \(\Delta_{t,j}=0\) for correct predictions,
\(U_{j,b}^{*}=e_{j,b}^{*}V_{j,b}^{*}\). The risk-estimation error
satisfies

\[
\begin{aligned}
|U_{j,b}^{*}-\widehat e_{j,b}\widehat V_j|
\leq{}&e_{j,b}^{*}|V_{j,b}^{*}-V_j^{\rm ref}|+e_{j,b}^{*}|V_j^{\rm ref}-\widehat V_j|+\widehat V_j|e_{j,b}^{*}-\widehat e_{j,b}|.
\end{aligned}
\]

The three terms represent context transfer, task-impact estimation, and
error-rate calibration. If \(\tau_{t,j}=\lambda H_{t,j}\) and

\[
|U_{j,b}^{*}-\widehat d_{t,j}|
<|\widehat d_{t,j}-\tau_{t,j}|,
\]

the estimated and bin-conditional risks make the same retention
decision. The gate is stable when its estimation error is smaller than
its distance from the threshold.

\begin{table}[!htbp]
\caption{Held-out checks of the two risk components. (a) Answer-NLL damage
conditioned on prediction errors for the original queries and independently
constructed unseen-query variants ($10^{-3}$ bits/answer). (b) Frozen entropy
lookup versus observed prediction error (\%).}
\label{tab:risk-component-checks}
\centering
\footnotesize
\begin{minipage}[t]{0.56\textwidth}
\centering
\textbf{(a) Error-conditioned damage}\par\vspace{2pt}
\setlength{\tabcolsep}{3.5pt}
\begin{tabular*}{\linewidth}{@{\extracolsep{\fill}}lrrrr@{}}
\hline
Subset & Coords. & Low-$V_j$ & \shortstack{Unseen\\query} & Random \\
\hline
Bottom 5\%  &  51 & \textbf{0.212} & 0.323 & 37.4 \\
Bottom 10\% & 102 & \textbf{0.221} & 0.375 & 38.8 \\
Bottom 15\% & 154 & \textbf{0.231} & 0.423 & 38.3 \\
Bottom 20\% & 205 & \textbf{0.232} & 0.565 & 39.8 \\
\hline
\end{tabular*}
\end{minipage}\hfill
\begin{minipage}[t]{0.41\textwidth}
\centering
\textbf{(b) Entropy and prediction error (\%)}\par\vspace{2pt}
\setlength{\tabcolsep}{3.5pt}
\renewcommand{\arraystretch}{1.2}
\begin{tabular*}{\linewidth}{@{\extracolsep{\fill}}lrrr@{}}
\hline
Entropy (bit) & Lookup & Observed & Gap (pp) \\
\hline
$[0,0.5)$   &  4.0 &  8.1 & $+4.1$ \\
$[0.5,1)$   &  9.5 & 16.0 & $+6.5$ \\
$[1,1.5)$   & 17.2 & 26.2 & $+9.1$ \\
$[1.5,2)$   & 27.1 & 35.1 & $+8.0$ \\
$[2,2.5)$   & 36.5 & 45.3 & $+8.8$ \\
$[2.5,3)$   & 44.8 & 52.5 & $+7.7$ \\
$[3,4)$     & 57.5 & 62.5 & $+5.0$ \\
$[4,5)$     & 76.3 & 78.8 & $+2.5$ \\
$[5,6)$     & 89.0 & 88.8 & $-0.2$ \\
\hline
\end{tabular*}
\end{minipage}
\end{table}
\FloatBarrier

\paragraph{The task-impact prior transfers across trajectories.}
On unseen trajectories, matched random sets cause $166$--$176\times$ as much
damage as the bottom $5\%$--$20\%$ of coordinates ranked by training-set $V_j$
(Table~\ref{tab:risk-component-checks}a). Conditioning on incorrect predictions
isolates error severity, supporting $V_j$ as a reusable task-impact prior.
For each evaluation question, we additionally construct a distinct unseen-query
variant. Damage remains $0.323$--$0.565$ across the four low-$V_j$ subsets,
far below matched random omission, indicating that the learned task-impact
prior remains relatively stable for rare or previously unseen query formulations.
A sufficient condition for this stability is given above in this section.

\paragraph{Entropy orders state-dependent prediction risk.}
Observed error rises monotonically from $8.1\%$ to $88.8\%$ across entropy
bins (Table~\ref{tab:risk-component-checks}b). The frozen lookup preserves this ordering
on held-out trajectories, supporting entropy as a signal for adapting the
error estimate $e_{t,j}$ to the current predictive context.

\section{Training-Data Construction and Evaluation-Overlap
Checks}\label{app:webarena-data}

\subsection{Unified collection
framework}\label{a.6.1-unified-collection-framework}

Web-related training data for AMA-Bench, WorldMemArena, LongMemEval-V2,
and WebQuest use WebArena as the common browser environment. Collection
follows its self-hosted sites, task configurations, browser state, and
execution logic. A local language model drives the agent, while a
lightweight adapter changes only the observation, action, and
trajectory-export interfaces. Each step is saved as a
state-action-next-state transition \((s_t,a_t,s_{t+1})\) with task/goal,
URL, accessibility tree, screenshot, element information, action, and
success outcome.

Both original and normalized semantic actions are retained, including
CLICK(target), TYPE(target,text), SCROLL(direction), GOTO(url), and
BACK(). These are action-class examples; the structured world-model
encoder uses the 17 action types defined above.

\subsection{Four training-data
sources}\label{a.6.2-four-training-data-sources}

AMA-Bench Web tasks use the WebArena agent loop with the full
accessibility tree and element-ID actions. Screenshots are synchronized
but are not required by the AMA-style interface. Successful and failed
trajectories are retained.

WorldMemArena agents primarily observe screenshots and use visual GUI
actions such as coordinate clicks, keyboard input, and scrolling.
Accessibility trees, URLs, and element information are logged in the
background. Long trajectories are segmented into sessions after
collection using page changes, subgoal completion, error feedback, or
clear state changes. Session boundaries, memory points, and QA are
post-processing annotations.

LongMemEval-V2 WebArena tasks use a BrowserGym/AgentLab-style interface
and store screenshots and accessibility trees together. Actions such as
click(bid), fill(bid,value), scrolling, navigation, and back are
normalized. Informative failures and multiple rollouts are retained to
cover errors and recovery, while episode boundaries and temporal order
are preserved.

WebQuest-style Trace QA stores complete screenshot sequences and marks
2--5 answer-bearing key frames. Questions require combining multiple
frames, such as for product comparison, counting, or attribute
filtering, and do not copy the original WebArena task goal. Items are
removed if one screenshot suffices, the answer is non-unique, or one
marked frame can be removed without losing the answer. The final sample
contains the screenshot sequence, question, and answer; accessibility
trees and actions are used only for generation, localization, and
verification.

\subsection{Deduplication}\label{a.6.3-deduplication}

Before collection, training and public evaluation tasks are lowercased,
non-semantic whitespace and punctuation differences are removed, and URL
and hostname forms are normalized. Task instructions and goals are then
matched exactly after normalization. Evaluation tasks that map to
WebArena tasks are excluded from the training pool.

When task IDs are unavailable, near-duplicate checks use the normalized
goal, starting URL, site or environment, and initial observation. Exact
or highly similar tasks are removed. A small random sample is manually
checked to verify that matched items refer to the same evaluation task,
that training goals do not copy evaluation goals, and that collection
trajectories do not copy evaluation trajectories. Training and
evaluation still share the website environment.

\subsection{MemEye synthetic training
data}\label{app:memeye-data}

\subsubsection{Visual structure and instance
separation}\label{a.6.4.1-visual-structure-and-instance-separation}

MemEye Health provides an evaluation set but no training split. We
therefore construct synthetic training data based on the structure of
the public benchmark, generating new virtual patients, screens,
values, and questions without reusing public evaluation images,
questions, or patient records. Dashboard
screens contain a title bar, KPI cards, a trend card, and three donut
cards; portal screens contain time-varying messages. The generator
samples new layouts, labels, values or text badges, donut labels and
colors, trend shapes, and value ranges. This construction covers the
evidence types and cross-screen comparisons required by evaluation, but
its task structure still reflects the public evaluation set. The MemEye
experiment therefore assesses memory under this class of screens and
questions; it does not establish zero-shot generalization to unseen
interface styles.

A parameterized PIL renderer creates a dashboard or portal on a
\(4096\times4096\) canvas and resizes it to \(1024\times1024\); trend
curves use 13 normalized control points. Structural parameters are saved
with each observation as ground truth for question generation and
retrieval supervision.

\subsubsection{Cross-session trajectories and
questions}\label{a.6.4.2-cross-session-trajectories-and-questions}

Each generated patient contains 8--12 dated sessions, including image
and text-only rounds. Metrics and portal guidance evolve across
sessions, and screen roles provide temporally distinguishable evidence.
A uniqueness check is applied to screen names and donut-card
fingerprints when answers and distractors are generated.

Seven question templates cover donut and layout identification,
cross-screen value comparison, trend shape, first-session content,
portal-guidance changes, and metric sequences. Clue-bearing rounds
provide multiple positive retrieval labels. Multiple-choice questions
use four option orderings and have open-ended counterparts; items with
conflicting answers or ambiguous
screen identities are removed.

\section{Persistent-Storage Accounting for
Baselines}\label{app:storage-accounting}

\subsection{Unified accounting
rules}\label{a.7.1-unified-accounting-rules}

To keep methods comparable, the main table counts logical persistent
content that must remain after memory construction to support future
queries. Under this accounting protocol, count serialized text,
structured fields, identifiers, timestamps, links, graph records, and
per-observation metadata that cannot be recovered losslessly from
counted content. If retrieval or answering reads media through a path,
ID, or resource node, count the media bytes; converted images are
counted in their converted form. A shared media asset is counted once.

If an image is used only during writing and later retrieval relies
entirely on saved text or structured memory, the original image may be
deleted and is excluded under this protocol. Regenerable embeddings and
retrieval indices, temporary caches, query-time working state, and
physical database-layout overhead are excluded. This convention compares
logical memory load rather than the physical disk usage of a particular
database implementation; physical deployment overhead should be reported
separately if it is relevant. Encoding, embedding, extraction,
retrieval, and reading parameters are functional model weights rather
than per-episode payload. MemoWM reports episode-specific storage in the
main table; the world-model cost is amortized over retained interactions
in Section A.4. The denominator is the number of retained observations,
and 1 kbit equals 1,000 bits.

MemoWM (w/o WM) uses the same observation encoder and OPQ interface but
stores every full quantized state and applicable auxiliary content,
without world-model prediction or state-dependent coordinate retention.

\subsection{Persistent content counted for each
baseline}\label{a.7.2-persistent-content-counted-for-each-baseline}

RAG and UniversalRAG count retained raw observations, media, text and
image records, identifiers, order, and required associations. A-Mem
counts retained MemoryNotes and their keywords, contexts, tags, links,
IDs, timestamps, scores, counters, and evolution fields. SimpleMem
counts MemoryEntry text, structured attributes, IDs, and metadata.
MemGPT counts Recall/Archival content and summaries needed by future
queries.

Omni-SM counts persistent text and visual memory units, retained visual
assets, and surviving events, entities, relations, and source links. M2A
counts raw media, raw messages, semantic records, captions, IDs,
timestamps, and evidence mappings, with shared images counted once.
ViLoMem counts retained logic records and textual visual descriptions,
and counts source images only when later access reads them. MIRIX counts
actual episodic, semantic, procedural, resource, knowledge-vault, and
optional core-memory records with IDs and metadata. AUGUSTUS counts
observation text and media, concept tags, graph nodes and edges, index
keys, and posting pointers; images remain counted when graph nodes allow
later access. MMFU-Single, when evaluated, follows the raw-content rule
used for UniversalRAG.

\subsection{Query-dependent memory
states}\label{a.7.3-query-dependent-memory-states}

MEM1 compact internal states and MemAgent rolling memories depend on the
current goal or question. The main table uses a reusable-memory
accounting: because future questions are unknown when a trajectory is
saved, it counts the original observations and media, actions, order,
and session metadata needed to construct states for different questions.
A state generated temporarily for one query is excluded unless it is
explicitly archived for later use. A single-query storage number may be
reported separately only when one known question is compressed and the
original trajectory is deleted.

\section{Physical Entropy-Coded Realization of the Reported Storage
Rate}\label{app:entropy-coding}

The main storage results use ideal conditional code lengths to provide an
implementation-independent measure of logical persistent memory. To verify that
these rates are realizable under a concrete coding implementation, we
additionally instantiate the MemoWM representation as an actual entropy-coded
bitstream on the same 817-transition WorldMemArena Web out-of-domain evaluation
set used in our rate analysis. The main paper reports an ideal
experience-specific rate of $4{,}323.84$ bits/state on this set.

We implement the retained-symbol stream using a range coder with 16-bit
cumulative-probability precision. At each transition, the writer and decoder
independently recompute the same predictive distributions, predictive
entropies, retention decisions, and modal predictions from the shared
reconstructed history. When a coordinate is retained, its true discrete value
is entropy-coded under the corresponding world-model predictive distribution.
When a coordinate is omitted, no symbol is written to the stream and the
decoder reconstructs it from the shared modal prediction. Importantly, no
explicit per-coordinate or per-state retention mask is stored.

The ideal-code accounting gives a rate of $4{,}323.84$ bits/state, while the
resulting serialized entropy-coded stream requires $4{,}388.70$ bits/state. The
physical implementation therefore adds $64.86$ bits/state, corresponding to
only $1.50\%$ overhead relative to the ideal-code estimate.

We further decode the complete bitstream independently from the stored initial
states, actions, shared model parameters, and coded payload. The decoder
reproduces $100.00\%$ of the writer-side reconstructed states exactly,
confirming that the retention decisions can be recomputed from shared
reconstructed history and do not require a hidden state-specific mask.

These results show that the ideal conditional code lengths used in the main
storage analysis closely approximate the size of an actual decodable
representation. In particular, replacing ideal code lengths with a physical
entropy-coded stream increases the measured rate by only $1.50\%$, while
leaving the reconstructed memory exactly unchanged. Thus, the storage
reduction reported in the main paper is not an artifact of ideal-code
accounting or an uncounted retention mask.

\section{Evaluation Settings}\label{app:evaluation-settings}

\textbf{WorldMemArena.} The Web subset uses a frozen Qwen3.5-9B as
Reader and LLM judge. Answers use greedy decoding, and the
benchmark-defined QA-C accuracy is reported.

\textbf{AMA-Bench.} The Web Task Execution subset follows the
benchmark-provided open-ended QA configuration with WebArena
trajectories. The base model is thinking-enabled Qwen3-32B. The
Qwen3-32B judge receives the question, reference answer, and prediction
under the corresponding evaluation prompt and returns a binary
correctness decision.

\textbf{LongMemEval-V2.} We use the Small setting, with histories from
100 trajectories in each of WebArena and ServiceNow. Following the
benchmark-provided context-collection protocol, trajectories are written
to memory, queried, and returned contexts are limited to 200K tokens.
Qwen3.5-9B serves as Reader and judge; the Reader uses thinking,
temperature 0.6, top-\(p=0.95\), and top-\(k=20\). Domain-specific
prompts and answer-parsing rules are retained; when a final boxed answer
is present, it is extracted before matching. Structured answers use the
corresponding matching rules, while trap and false-premise questions use
the benchmark semantic-scoring prompt.

\textbf{WebQuest.} We follow the benchmark Trace QA evaluation and
report Relaxed Accuracy (RA). Numeric values are extracted from
predictions and reference answers and compared with tolerance
\(\pm0.05\); integers and strings use normalized exact matching. A
prediction is correct if it matches any acceptable reference answer.

\textbf{Storage and reconstruction diagnostics.} One kbit is 1,000 bits.
Storage rate is persistent bits per episode divided by the number of
observations. Controlled reconstruction distortion is the absolute
change in teacher-forced NLL relative to complete quantized memory,
reported in bits per answer.

\end{document}